\pdfoutput=1

\documentclass[11pt]{article}

\usepackage{EMNLP2023}  

\usepackage{times}
\usepackage{latexsym}

\usepackage[T1]{fontenc}

\usepackage[utf8]{inputenc}

\usepackage{microtype}

\usepackage{inconsolata}
\usepackage{tabularx}

\usepackage{booktabs}
\usepackage{multirow}
\usepackage{adjustbox}
\usepackage{amssymb}
\usepackage{pifont}
\usepackage[linewidth=0.5pt]{mdframed}

\usepackage{caption}
\usepackage[font={small}]{caption}
\usepackage{etoolbox}
\usepackage{hhline}
\usepackage{tikz}
\usepackage{collcell}
\usepackage{enumitem}

\usepackage{amsmath}
\setlist[itemize]{noitemsep}
\usepackage{float}
\usepackage{arydshln}

\DeclareMathOperator*{\argmax}{arg\,max}

\newcommand{\comment}[1]{}

\title{Introducing the NewsPaLM MBR and QE Dataset:

LLM-Generated High-Quality Parallel Data Outperforms Traditional Web-Crawled Data}

\author{
\textbf{Mara Finkelstein} \quad
\textbf{David Vilar} \quad
\textbf{Markus Freitag} \\
Google \\
  \texttt{\{marafin,vilar,freitag\}@google.com}
}

\begin{document}
\maketitle

\begin{abstract}
Recent research in neural machine translation (NMT) has shown that training on high-quality machine-generated data can outperform training on human-generated data.
This work accompanies the first-ever release of a LLM-generated, MBR-decoded and QE-reranked dataset with both sentence-level and multi-sentence examples.\footnote{The dataset can be found at \url{https://github.com/google-research/google-research/tree/master/newspalm\_mbr\_qe}, and is released under an Apache 2.0 license.}
We perform extensive experiments to demonstrate the quality of our dataset in terms of its downstream impact on NMT model performance.
We find that training from scratch on our (machine-generated) dataset outperforms training on the (web-crawled) WMT'23 training dataset (which is 300 times larger), and also outperforms training on the top-quality subset of the WMT'23 training dataset.
We also find that performing self-distillation by finetuning the LLM which generated this dataset outperforms the LLM's strong few-shot baseline.
These findings corroborate the quality of our dataset, and demonstrate the value of high-quality machine-generated data in improving performance of NMT models.

\end{abstract}

\section{Introduction}
\label{sec:intro}

With the advent of large language models (LLMs), machine translation (MT) quality has improved dramatically \cite{kocmi-etal-2023-findings,kocmi2024preliminarywmt24rankinggeneral}, and performance tends to scale with model size \cite{geminiteam2024geminifamilyhighlycapable}.
While LLMs are now state-of-the-art translators, they are often impractical to use or serve, especially in high-traffic and/or resource-constrained settings.
Thus, development of smaller, but still highly performant, MT models remains an active area of research. Recent work has shown that distillation of LLM translation quality, while requiring an expensive data generation process, is an effective approach \citep{li-etal-2024-mt}. In this work, we introduce a new LLM-generated dataset called \texttt{NewsPaLM}, which we make freely available.

In addition to the size of the teacher model, another key determinant of the quality of machine-generated translation data is the decoding method used. While beam search and greedy decoding are the most common decoding methods used for NMT, \citet{eikema2020map} showed that maximum \textit{a posteriori} (MAP) decoding methods are suboptimal, and instead proposed Minimum Bayes Risk (MBR) decoding.
Unlike MAP decoding, MBR decoding does not aim to produce the translation with the highest estimated model probability.
Instead, it chooses the translation that is estimated to have the highest quality with respect to a utility metric.
A follow-up study by \citet{freitag2022high} showed that MBR decoding with neural utility metrics significantly outperforms beam search decoding, according to expert-based human evaluation.

The main drawback of MBR decoding is its high computational cost.
In particular, the algorithm requires that, for every input query, a large number $n$ of candidates be generated from the model, and then an (expensive) scoring function be computed on every pair of distinct candidates $(n_i, n_j)$, for a total of $O(n^2)$ computations.
QE reranking \cite{fernandes-etal-2022-quality} is a more efficient alternative to MBR decoding. This decoding method instead reranks the candidate model predictions using a neural quality estimation (QE) metric, and requires only $O(n)$ computations.

\citet{finkelstein2023mbr} showed that finetuning NMT models on MBR-decoded and QE-reranked datasets is an effective technique for distillation (while finetuning on beam search-decoded datasets is not) and that, given a LLM teacher, MBR and QE distillation can outperform finetuning on human-generated references.

In this work, we generate sentence-level parallel data using MBR decoding and multi-sentence parallel data using QE reranking. In addition to detailing our dataset creation method, we also perform extensive experiments to demonstrate the quality of our dataset in terms of its downstream impact on NMT model performance.

Our contributions can be summarized as follows:
\begin{itemize}
  \item We release our LLM-generated, sentence-level and multi-sentence, MBR and QE translation dataset.
  \item We demonstrate that our dataset is high-quality, by using it to train NMT models from scratch and comparing performance against baselines using human-generated parallel data. This is the first work to pretrain NMT models on MBR and QE data.
  \item We show that training on our dataset outperforms training on the web-crawled WMT'23 training dataset (which is 300 times larger than ours). Moreover, our dataset also outperforms (by an even larger margin) when compared against quality-based filtering of the WMT'23 dataset to match the size of our dataset.
  \item We also demonstrate our dataset's quality by performing self-distillation (using the PaLM-2 LLM from which this data was generated), and show that this outperforms the LLM's strong few-shot baseline. To our knowledge, this is the first work to investigate MBR finetuning a LLM.
  \item We investigate the effect of sentence-level versus multi-sentence MBR and QE training data on NMT model performance as a function of sequence length, and investigate the tradeoff between dataset size and model quality, during both pretraining and finetuning.
\end{itemize}

\section{\texttt{NewsPaLM} Dataset}
\label{sec:datasets}

This paper accompanies a dataset release of sentence-level and multi-sentence English-German and German-English parallel data, generated from the (monolingual) Newscrawl corpus as made available for the WMT evaluation campaigns\footnote{The Newscrawl data was downloaded from \url{https://data.statmt.org/news-crawl/}.} using the \textit{PaLM-2 Bison} LLM~\citep{anil2023palm}.
We detail below the steps to create this dataset, which we call \texttt{NewsPaLM}.

The dataset construction process consisted of four steps, as described in the following sections.

\subsection{Source-side Data Collection: Newscrawl}
\label{sec:newscrawl}
To construct the English and German source-side datasets, we first collected all Newscrawl data from 2007 to 2022, released as part of the WMT'23 Machine Translation Shared Task \cite{kocmi-etal-2023-findings}.
This is a large corpus of crawled news, with about 398 million and 507 million lines for English and German, respectively. For both of these languages, document-split versions of the dataset (with document boundaries intact) are available.

We collected both the sentence-level and document-level versions of the datasets. Basic preprocessing had already been applied to the sentence-level version, including removing lines with no
ASCII letters and deduplication. This preprocessing was not applied to the document-level version. We performed minimal additional cleaning to fix incorrectly encoded characters.

\subsection{Construction of ``Blobs''}
\label{sec:blobs}
We used the document-split versions of the datasets to construct multi-sentence (i.e. ``blob-level'') examples. We refer to these examples as blobs, rather than paragraphs, since they do not respect paragraph boundaries but, rather, simply represent the concatenation of contiguous sentences up to a maximum length. In particular, we joined headlines using the separator ``\texttt{\textbackslash n\textbackslash n}'', and otherwise joined sentences with spaces, up to a maximum length of 512 tokens (using the PaLM-2 tokenizer;~\citet{anil2023palm}). The blobs respect document boundaries, each blob contains only complete sentences (no sentence fragments), and each blob may or may not contain a headline (depending on where in the document the blob comes from).

\subsection{Cluster-Based Text Selection}
\label{sec:maat}
The size of the Newscrawl full dataset and the high computational cost of the decoding techniques (\S\ref{sec:mbr_qe_decoding}) makes it impractical to process all the available data.
In order to reduce the size of the data, while at the same time ensuring diversity in the samples, we follow a clustering-based sample selection approach.
As a first step, we embed the source side of the data using XLM-RoBERTa \cite{conneau-etal-2020-unsupervised}.
We then apply Recursive Agglomerative Clustering~(RAC) \cite{sumengen2021scaling}, an unsupervised clustering algorithm which is an efficient extension of Hierarchical Agglomerative Clustering.
These algorithms are initialized by defining a set of clusters, each containing a single point from the original data points.
The guiding principle is to iteratively merge the two clusters which are closest to each other, until some stopping criterion is met, e.g.~a maximum distance between the clusters to be merged.
Note that this algorithm requires the number of clusters to be chosen as a hyperparameter, unlike other clustering algorithms like $k$-means which have the advantage that the number of clusters is defined by the algorithm itself.
We selected the number of clusters shown in Table~\ref{tab:number_of_clusters} for each of the data sets.

\begin{table}
  \begin{tabular}{lcc}
    \toprule
                   & \bf EN $\rightarrow$ DE & \bf DE $\rightarrow$ EN \\
     \midrule
    Sentence-level & 3,287                   & 3,264                   \\
    Blob-level     & 3,826                   & 4,017                   \\
    \bottomrule
\end{tabular}
\caption{Number of defined clusters per dataset.}
\label{tab:number_of_clusters}
\end{table}

Once the clusters have been defined, we sample uniformly from them.
In this way, we ensure that the diversity of the original dataset is maintained in the reduced sample.

\subsection{MBR Decoding and QE Reranking}
\label{sec:mbr_qe_decoding}
The preceding steps handle preparation of source-side data. To generate the target-side data from these sources, we used the \textit{PaLM-2 Bison} LLM~\citep{anil2023palm}, 5-shot prompted with ICL examples from the newstest2021 test set~\citep{akhbardeh-etal-2021-findings}. Note that unlike previous work which also used PaLM-2 to generate translation data for distillation \citep{finkelstein2023mbr}, here we do not finetune on the translation task prior to data generation.

A key component of our data generation process is the decoding method. We generated the sentence-level data using MBR decoding and the blob-level data using QE reranking. Both MBR decoding and QE reranking can be decomposed into two steps: candidate list generation (Section \S~\ref{sec:mbr_cand_list}) and scoring (Section \S~\ref{sec:mbr_scoring}).

\subsubsection{Candidate List Generation}
\label{sec:mbr_cand_list}
The first step in the decoding process is to generate a list of candidate model outputs, given a source segment. In this work, we used a candidate size of 512 and generated candidate translations using epsilon sampling \citep{hewitt2022truncation} with $\varepsilon=0.02$, which was shown to be the best sampling method for MBR decoding in \citet{freitag2023epsilon}.

\subsubsection{MBR and QE scoring}
\label{sec:mbr_scoring}
Next, the best output is chosen based on a utility function.
This step is where MBR decoding and QE reranking diverge.
For MBR decoding, we use a reference-based utility metric $u_{mbr}(h,r)$, which estimates the quality of a candidate translation $h$ conditioned on a reference translation $r$.
Formally, given a set of hypotheses $\mathcal{H}$, the Minimum Bayes Risk (MBR) translation $h^{mbr}$ is selected from the candidates in $\mathcal{H}$ according to
\[h^{mbr} = \argmax_{h \in \mathcal{H}} \frac{1}{|\mathcal{H}|} \sum_{y \in \mathcal{H}} u_{mbr}(h, y).\]
For QE reranking, on the other hand, we use a reference-free (QE) utility metric $u_{qe}(h, s)$, which estimates the quality of a candidate translation $h$ conditioned on the source $s$, rather than on the reference.
We select the best QE translation $h^{qe}$ of the source $s$ from the candidates in $\mathcal{H}$ as
\[h^{qe} = \argmax_{h \in \mathcal{H}} u_{qe}(h, s)\]
In this work, we used the \textit{BLEURT} \citep{sellam2020bleurt} utility metric for MBR decoding and the \textit{MetricX-QE} \citep{juraska-etal-2023-metricx} utility metric for QE reranking. Note that the maximum context length for \textit{BLEURT} (candidate and reference combined) is 512, while for \textit{MetricX-QE} (candidate and source combined), it is 1024. Given that the blob-level source-side data alone can contain up to 512 tokens, we could not use \textit{BLEURT} as the utility function for this data. MBR decoding with \textit{MetricX} is prohibitively expensive, hence our decision to perform QE reranking instead.

As a baseline against which to compare data generated using these state-of-the-art decoding methods, we also created accompanying sentence-level and blob-level datasets from the same source-side data using greedy decoding.

\subsection{Dataset Statistics}

Here we briefly present basic statistics about the four datasets we created (sentence-level and blob-level versions, for the en$\rightarrow$de and de$\rightarrow$en language pairs).
See Appendix \ref{appendix:a} for additional dataset statistics.

Table~\ref{tab:dataset_sizes_mbr_qe} shows the size (in number of examples) of each dataset.
Note that each dataset has about 800 thousand to one million examples.
\begin{table}
    \centering
    \begin{adjustbox}{width=\columnwidth}
\begin{tabular}{ccc}
    \toprule
     & \bf EN $\rightarrow$ DE & \bf DE $\rightarrow$ EN\\
    \midrule
    \textsc{MBR sent-level} & 998,435 & 1,022,344 \\
    \textsc{QE blob-level} & 925,829 & 769,028 \\
    \bottomrule
\end{tabular}
    \end{adjustbox}
    \caption{Number of examples per dataset.}
    \label{tab:dataset_sizes_mbr_qe}
\end{table}

\begin{figure}
    \centering
    \includegraphics[width=\columnwidth]{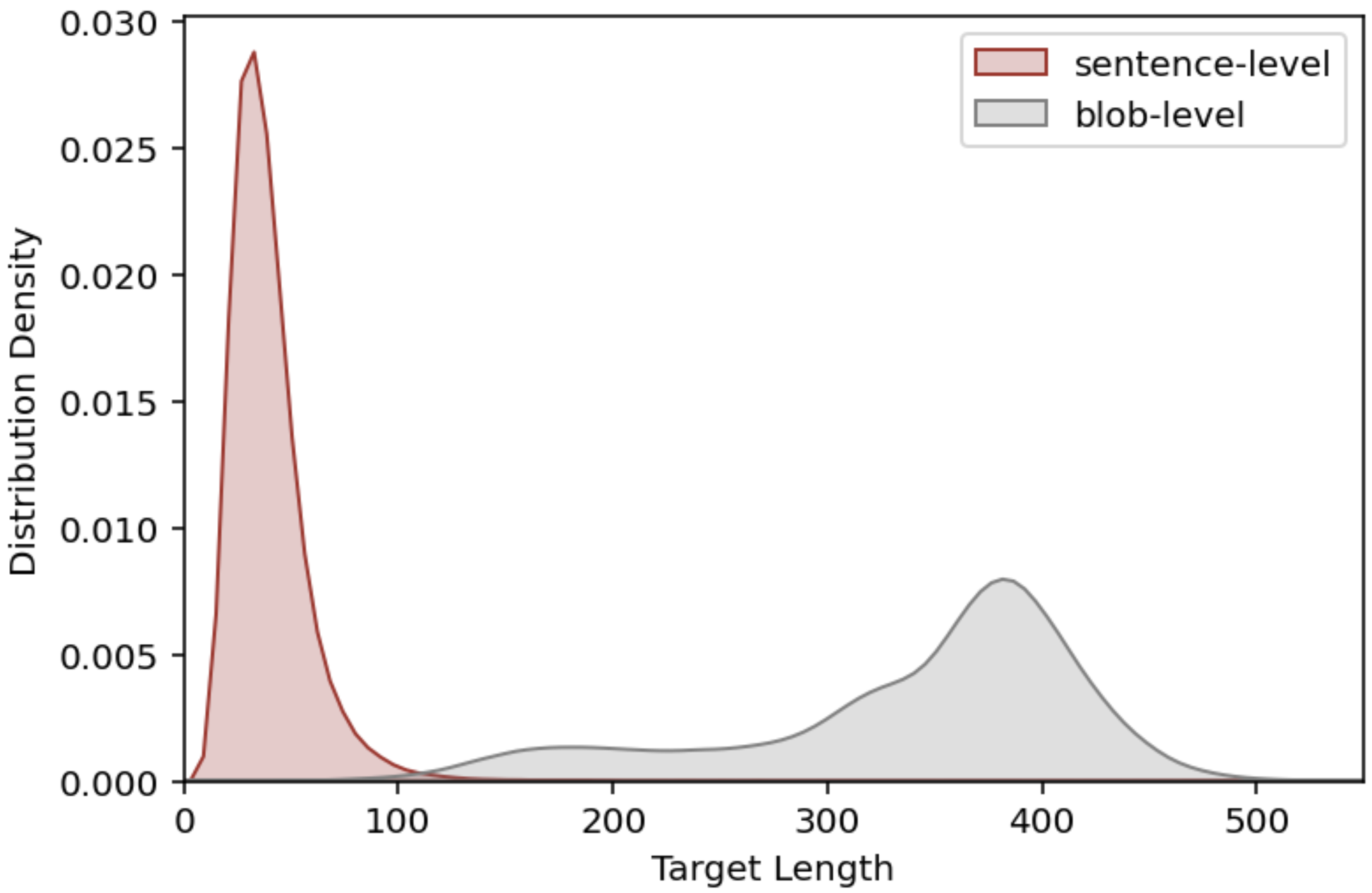}
    \caption{Distribution of English-German MBR sentence-level versus QE blob-level target lengths (computed using the \texttt{Moses} tokenizer).}
    \label{fig:dataset_size_distributions}
\end{figure}

Table~\ref{tab:src_tgt_lengths} shows the average length of source and target examples (in number of tokens, as defined by the \texttt{Moses} tokenizer) per dataset.
For English-German, the blob-level examples are about ten times longer than the sentence-level examples, while for German-English, they are about four times longer.
Figure~\ref{fig:dataset_size_distributions} shows the distribution of target example lengths for English-German.
Note that the blob-level data distribution is shifted to the right of the sentence-level data distribution, as expected.
\begin{table}
    \centering
\begin{tabular}{lccc}
    \toprule
     & & \bf Source & \bf Target \\
    \midrule
    \multirow{2}{*}{\textsc{EN $\rightarrow$ DE}} &
    \textsc{Sentences} & 37.5 & 39.8 \\
    & \textsc{Blobs} & 364.5 & 339.8 \\
    \midrule
    \multirow{2}{*}{\textsc{DE $\rightarrow$ EN}} &
    \textsc{Sentences} & 77.3 & 88.3 \\
    & \textsc{Blobs} & 288.4 & 323.4 \\
    \bottomrule
\end{tabular}
    \caption{Average source and target lengths per dataset, computed using the \texttt{Moses} tokenizer.}
    \label{tab:src_tgt_lengths}
\end{table}

\section{Experimental Setup}
\label{sec:setup}

We perform a series of pretraining and finetuning experiments to validate the quality of our \texttt{NewsPaLM} dataset, and to contextualize its performance with respect to a much larger dataset of human-generated data.
All of our experiments are performed on both English-German (en$\rightarrow$de) and German-English (de$\rightarrow$en).

\subsection{Datasets}
\subsubsection{Training Data}
\label{sec:training_data}
As a baseline against which to compare our \texttt{NewsPaLM} dataset, we use the parallel WMT'23 training data \citep{kocmi2023findings}, which consists of 296 million sentence-level examples. A subset of this data (consisting of about 3 million sentences, from Europarl, News Commentary, and Rapid documents) contains document boundaries, which we use to construct blob-level examples using a procedure similar to the blob-level dataset creation process described in Section \S\ref{sec:blobs}.
That is, we partition the sentences into contiguous blocks, each of which has a total number of tokens up to a token limit of 512 (for each of source and target).
In our experiments, this WMT'23 data is only used for pretraining.

The remainder of our pretraining and finetuning data comes from our (machine-generated) \texttt{NewsPaLM} dataset, described in Section \S\ref{sec:datasets}. As an additional baseline, we compare the MBR-decoded and QE-reranked versions of this dataset against the greedy-decoded version. Note that for both language pairs (en$\rightarrow$de and de$\rightarrow$en), the sentence-level and blob-level \texttt{NewsPaLM} data combined contains less than 2 million examples (Table~\ref{tab:dataset_sizes_mbr_qe}).

\subsubsection{Development and Test Sets}
For both language pairs, we use the sentence-level and paragraph-level versions of the newstest2021 test set \citep{farhad2021findings}, as well as the (sentence-level) generalMT2022 test set \citep{kocmi-etal-2022-findings}, as our development sets for checkpoint picking.
We report all results on the WMT'23 \citep{kocmi2023findings} and WMT'24 \citep{kocmi2024preliminary} test sets.
Note that the WMT'23 and WMT'24 en$\rightarrow$de test sets are paragraph-level.

\subsection{Models}
\label{sec:setup_models}
For both language pairs (en$\rightarrow$de and de$\rightarrow$en), we use a 602 million parameter Transformer encoder-decoder architecture, implemented in \textit{Pax}\footnote{\url{https://github.com/google/paxml}}.
The model has 8 encoder and 8 decoder layers (rather than 6), but otherwise is similar to the \textit{transformer-big} setting in \citet{vaswani2017attention}, with model dimension of 1024, hidden dimension of 8192, and 16 multi-attention heads.
We train without label smoothing.
For each language pair, we use a bilingual vocabulary of 32k subword units trained on the WMT'23 training dataset \citep{kocmi2023findings}.
The best (base and incremental) checkpoints were chosen to maximize \textit{BLEURT}~\citep{sellam2020bleurt} on the development set.

We also experiment with self-distillation of the \textit{PaLM-2 Bison}~\citep{anil2023palm} LLM, which is the model used to generate our datasets (see Section \S\ref{sec:mbr_qe_decoding}).
We compare self-distillation (finetuning) against 5-shot prompting of this model (using the same ICL examples as during \texttt{NewsPaLM} dataset generation).

\subsection{Evaluation}
We evaluate our models on four automatic metrics:  \textit{MetricX} \citep{juraska-etal-2023-metricx}, \textit{Comet20} \citep{rei2020comet}, \textit{Comet22} \citep{rei2022comet}, and \textit{BLEURT} \citep{sellam2020bleurt}. Note that for \textit{MetricX}, lower scores are better, while for the remaining metrics, higher scores are better.
Since the MBR data is generated using \textit{BLEURT} as the utility function, and the QE data is generated using \textit{MetricX}, the MBR-finetuned models may overfit to the \textit{BLEURT} metric, while the QE-finetuned models may overfit to the \textit{MetricX} metric. Thus, we primarily depend on the \textit{Comet*} metrics to measure model quality.

\section{Results}
\label{sec:results}

\subsection{Pretraining}
\label{sec:results_pretraining}

We first experiment with training bilingual (en$\rightarrow$de and de$\rightarrow$en) encoder-decoder translation models (as described in \S\ref{sec:setup_models}) from scratch, to compare our \texttt{NewsPaLM} dataset (as described in \S\ref{sec:training_data}) against the WMT'23 training dataset.
As shown in Table~\ref{tab:pretraining}, \textbf{pretraining on the \texttt{NewsPaLM} QE blob-level dataset} (which contains less than one million examples; Table~\ref{tab:dataset_sizes_mbr_qe}) \textbf{outperforms pretraining on the entire WMT'23 training dataset}, which is more than 300 times larger.
The \texttt{NewsPaLM} QE dataset achieves a \textit{Comet22} score of 80.62 on the English-German WMT'23 test set (row 2c), while the WMT'23 training dataset achieves a score of 78.79 (row 1a).

\begin{figure}
    \centering
    \includegraphics[width=\columnwidth]{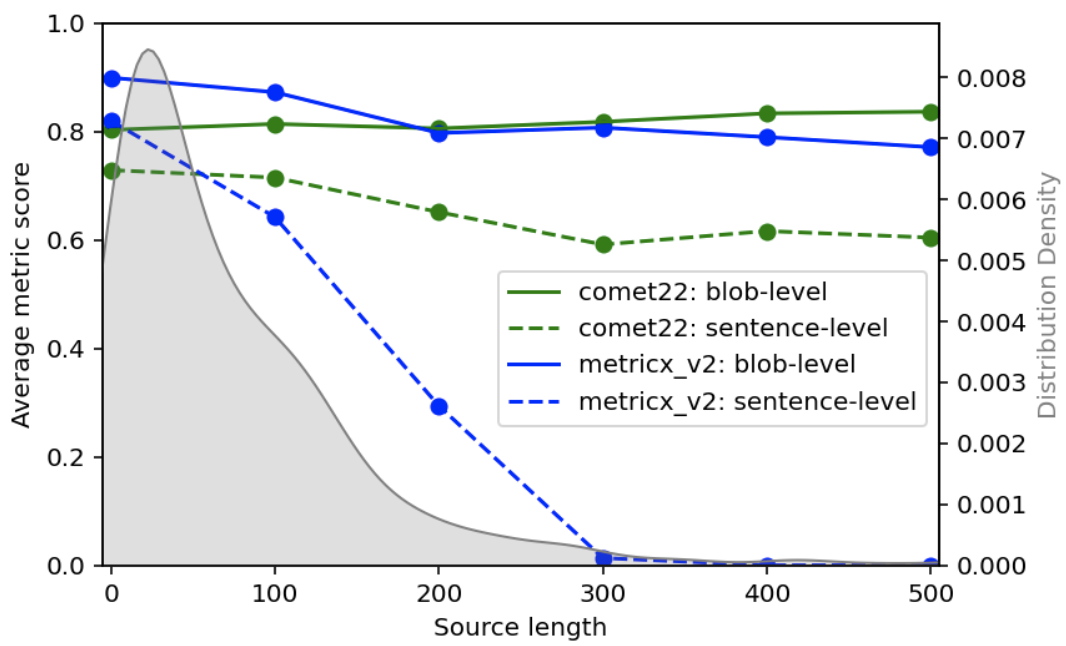}
    \caption{Comparison of pretraining performance on \texttt{NewsPaLM} MBR sentence-level dataset versus \texttt{NewsPaLM} QE blob-level dataset, bucketed by source length (WMT'23 en$\rightarrow$de test set). Note that performance of the model trained on the blob-level data is stable across segment lengths, while performance of the model trained on the sentence-level data declines as segment length increases (according to both \textit{MetricX} and \textit{Comet22} metrics).}
    \label{fig:results_bucketed}
\end{figure}

Note that \textbf{training on the MBR sentence-level data} (row 2b) \textbf{underperforms training on the QE blob-level data} (row 2c). As shown in Figure~\ref{fig:results_bucketed}, this is mostly due to a large drop in performance on longer sequence lengths. Thus, \textbf{exposure to multi-sentence data during training is essential to perform well on paragraph-level test sets}. Also note that during pretraining, we see no additional gains from mixing in the MBR sentence-level data relative to using the QE blob-level data only (rows 2c versus 2d). 

We also experiment with pretraining on the greedy-decoded version of our \texttt{NewsPaLM} dataset, to compare against pretraining on the MBR-decoded and QE-reranked versions.
Interestingly, the former (pretraining on the greedy-decoded data) outperforms the latter (pretraining on the MBR-decoded and QE-reranked versions), as shown in rows 2a versus 2d in Table~\ref{tab:pretraining}.
Based on manual inspection of examples, we hypothesize that the MBR-decoded and QE-reranked data is more free-style and harder for the model to learn than the greedy-decoded data.
This is illustrated in Table~\ref{tab:greedyVsMBR} in the Appendix.
If this were the case, the model would perform better by first learning the "easier" data (during pretraining), then adapting to the more free-style data during finetuning.
We test this hypothesis by comparing two model training curricula: For the first, we pretrain on the greedy-decoded data and finetune on the MBR-decoded and QE-reranked data.
For the second, we do the opposite: pretraining on the MBR-decoded and QE-reranked data and finetuning on the greedy decoded data.
As we hypothesized, the \textbf{former model training curriculum (MBR and QE finetuning from the greedy-pretrained checkpoint) performed better} (Table~\ref{tab:mbr-greedy}).

We have seen that pretraining on a small and clean, synthetically-produced dataset (\texttt{NewsPaLM}) can outperform finetuning on a large and noisy, human-generated one (WMT'23 training dataset). However, previous work such as \citet{peter2023there} has shown that MT model performance can be boosted by selecting a high-quality subset of a large and noisy training corpus, using data selection techniques such as QE filtering. Thus, we perform QE filtering (using the \textit{BLEURT-QE} metric, as in \citet{peter2023there}) to select the highest-quality examples in the WMT'23 (sentence-level) training dataset, while reducing its size to exactly match that of our (sentence-level) \texttt{NewsPaLM} dataset (of about one million examples). As shown in row 1b versus row 2b in Table~\ref{tab:pretraining}, \textbf{training on the QE-filtered WMT'23 dataset substantially underperforms training on our MBR-decoded \texttt{NewsPaLM} dataset (of the same size)}, and also underperforms training on the full WMT'23 dataset. Note that this result does not contradict previous work showing the benefit of data filtering, since previous work did not reduce the dataset to such a small fraction (0.3\%) of the original size. Thus, our \texttt{NewsPaLM} dataset is highly efficient (which is one indicator of its quality), and its efficiency cannot be matched be selecting a high-quality subset of a large, noisy corpus.

\begin{table*}[t]
    \centering
    \begin{adjustbox}{width=\textwidth}
\begin{tabular}{llcc}
\toprule
& \bf Model & \bf MetricX $\downarrow$ & \bf COMET22 $\uparrow$ \\
\midrule
\multirow{6}{*}{\textit{en$\rightarrow$de}}
& 1a) WMT'23 (all) & 4.20 &78.79 \\
& 1b) WMT'23 (sentence-level, \textit{BLEURT-QE} filtered) & 16.69 & 43.18 \\
\cmidrule{2-4}
& 2a) Greedy sentence-level + blob-level (9:1) & \bf 2.60 & \bf 81.67 \\
& 2b) MBR sentence-level & 6.39 & 72.05 \\
& 2c) QE blob-level & 2.82 & 80.62 \\
& 2d) MBR sentence-level + QE blob-level (9:1) & 2.99 & 79.68 \\
\midrule[1.5pt]
\multirow{6}{*}{\textit{de$\rightarrow$en}}
& 1a) WMT'23 (all) & 5.55 & 82.41 \\
& 1b) WMT'23 (sentence-level, \textit{BLEURT-QE} filtered) & 14.80 & 57.27 \\
\cmidrule{2-4}
& 2a) Greedy sentence-level + blob-level (9:1) & \bf 3.47 & \bf 83.30 \\
& 2b) MBR sentence-level & 4.97 & 80.55 \\
& 2c) QE blob-level & 4.01 & 82.33 \\
& 2d) MBR sentence-level + QE blob-level (9:1) & 3.95 & 82.02 \\
\bottomrule
\end{tabular}
\end{adjustbox}
\caption{Pretraining performance (WMT'23 test set).}
\label{tab:pretraining}
\end{table*}
\begin{table*}[t]
    \centering
    \begin{adjustbox}{width=\textwidth}
\begin{tabular}{llcccc}
\toprule
& \bf Model & \bf MetricX $\downarrow$ & \bf COMET22 $\uparrow$ \\
\midrule
\multirow{2}{*}{\textit{en$\rightarrow$de}}
& 1a) MBR + QE finetuning (from greedy-pretrained ckpt) & \bf 2.11 & \bf 82.78 \\
& 1b) Greedy finetuning (from MBR + QE-pretrained ckpt) & 2.63 & 81.48 \\
\midrule
\multirow{2}{*}{\textit{de$\rightarrow$en}}
& 1a) MBR + QE finetuning (from greedy-pretrained ckpt) & \bf 3.10 & \bf 84.05 \\
& 1b) Greedy finetuning (from MBR + QE-pretrained ckpt) & 3.60 & 83.08 \\
\bottomrule
\end{tabular}
\end{adjustbox}
\caption{Comparison of pretraining on \texttt{NewsPaLM} greedy data, then finetuning on \texttt{NewsPaLM} MBR and QE data, versus vice-versa (WMT'23 test set).}
\label{tab:mbr-greedy}
\end{table*}

\subsection{Finetuning}
\label{sec:results_finetuning}
Next, we experiment with how the different variants of our \texttt{NewsPaLM} dataset (and mixtures thereof) behave during finetuning (and whether this behavior differs from that observed during pretraining). Unless otherwise indicated, we initialize from the checkpoint pretrained on the WMT'23 training data (row 1a in Table~\ref{tab:pretraining}). We report en$\rightarrow$de and de$\rightarrow$en results on the WMT'23 test set in Table~\ref{tab:finetuning}, and refer the reader to Table~\ref{tab:wmt24-ende} in Appendix \ref{appendix:b} for pretraining and finetuning results on the WMT'24 en$\rightarrow$de test set.

\begin{table*}[t]
    \centering
    \begin{adjustbox}{width=\textwidth}
\begin{tabular}{llcc}
\toprule
& \bf Model & \bf MetricX $\downarrow$ & \bf COMET22 $\uparrow$ \\
\midrule
\multirow{8}{*}{\textit{en$\rightarrow$de}}
& 1a) Greedy sentence-level + blob-level (9:1) & 2.59 & 81.49 \\
& 1b) MBR sentence-level & 2.30 & \bf 82.69 \\
& 1c) QE blob-level & 2.45 & 81.83 \\
& 1d) MBR sentence-level + QE blob-level (9:1) & \bf 2.26 & 82.52 \\
\cmidrule{2-4}
& 2a) PaLM-2 five-shot (no finetuning) & 1.62 & 84.54 \\
& 2b) PaLM-2 MBR sentence-level & \bf 1.14 & \bf 85.64 \\
& 2c) PaLM-2 QE blob-level & 1.47 & 84.77 \\
& 2d) PaLM-2 MBR sentence-level + QE blob-level (9:1) & 1.17 & 85.54 \\
\midrule[1.5pt]
\multirow{8}{*}{\textit{de$\rightarrow$en}}
& 1a) Greedy sentence-level + blob-level (9:1) & 3.12 & 84.14 \\
& 1b) MBR sentence-level & 2.91 & \bf 84.57 \\
& 1c) QE blob-level & 2.99 & 84.27 \\
& 1d) MBR sentence-level + QE blob-level (9:1) & \bf 2.82 & 84.53 \\
\cmidrule{2-4}
& 2a) PaLM-2 five-shot (no finetuning) & 2.25 & 85.36 \\
& 2b) PaLM-2 MBR sentence-level & \bf 1.91 & \bf 86.26 \\
& 2c) PaLM-2 QE blob-level & 2.03 & 85.81 \\
& 2d) PaLM-2 MBR sentence-level + QE blob-level (9:1) & 1.92 & 86.24 \\
\bottomrule
\end{tabular}
\end{adjustbox}
\caption{Finetuning performance (WMT'23 test set). Unless otherwise indicated, performance is reported for the encoder-decoder model. For finetuning, this model was initialized from the checkpoint pretrained on the full WMT'23 training dataset (row 1a in Table~\ref{tab:pretraining}). Results for \textit{PaLM-2 Bison} few-shot prompting versus self-distillation using \texttt{NewsPaLM} MBR and QE data are reported in rows 2a-d.}
\label{tab:finetuning}
\end{table*}

As shown in Table~\ref{tab:finetuning}, \textbf{MBR and QE finetuning} (row 1d) \textbf{outperforms greedy finetuning} (row 1a), using the same mixture proportions (9:1) for the sentence-level and blob-level data. As shown in Table~\ref{tab:mbr-greedy} and discussed in \S\ref{sec:results_pretraining}, MBR and QE finetuning from the greedy-pretrained checkpoint outperforms greedy finetuning from the MBR and QE-pretrained checkpoint as well. Also, note that for en$\rightarrow$de, MBR and QE finetuning from the checkpoint pretrained on the WMT'23 training data (row 1d in Table~\ref{tab:finetuning}) slightly underperforms initializing from the checkpoint pretrained on the greedy \texttt{NewsPaLM} dataset (row 1a in Table~\ref{tab:mbr-greedy}) according to the WMT'23 test set, but the opposite is the case according to the WMT'24 test set (see Tables~\ref{tab:wmt24-ende} and \ref{tab:wmt24-mbr-greedy} in Appendix~\ref{appendix:b}) and based on the de$\rightarrow$en results on the WMT'23 test set.

Unlike during pretraining, \textbf{finetuning on the MBR sentence-level data outperforms finetuning on the QE blob-level data} (rows 1b versus 1c in Table~\ref{tab:finetuning}), and we see no additional gains from mixing in the QE blob-level data relative to using the MBR sentence-level data only (rows 1b versus 1d). We hypothesize that the model learns to use long context (from the blob-level data) during pretraining, and it doesn't forget during finetuning, so blob-level data is less important during this stage.

We also experiment with finetuning the \textit{PaLM-2 Bison}~\citep{anil2023palm} LLM (as described in \S\ref{sec:setup_models}), which is the teacher model used to generate our \texttt{NewsPaLM} dataset. As shown in Table~\ref{tab:finetuning}, \textbf{self-distillation via MBR (and QE) finetuning does indeed improve performance over the LLM's strong few-shot baseline} (rows 2a vs 2b). As with the encoder-decoder model, finetuning \textit{PaLM-2 Bison} on the MBR data outperforms finetuning on the QE data (and outperforms finetuning on a mixture of the MBR and QE data). The improvement in performance of \textit{PaLM-2 Bison} due to MBR finetuning is observed across all source length buckets (Figure~\ref{fig:results_ulm_bucketed} in Appendix~\ref{appendix:b}) and all domains in the WMT'23 and WMT'24 test sets (Tables~\ref{tab:wmt23-ende-per-domain} and \ref{tab:wmt24-ende-per-domain} in Appendix~\ref{appendix:b}), despite the MBR data being sentence-level only and coming primarily from the news domain. MBR finetuning the \textit{PaLM-2 Bison} model also outperforms MBR finetuning the much smaller encoder-decoder student (rows 1b vs 2b in Table~\ref{tab:finetuning}), as expected.

\subsection{Ablations}
\subsubsection{Effect of Dataset Size}
\label{sec:ablation_scaling}
Given the expense of creating LLM-generated, MBR-decoded datasets such as the ones presented in this work, we investigate how model performance scales with dataset size during both pretraining and finetuning. We randomly sample 25\% of the MBR-decoded \texttt{NewsPaLM} dataset (for both en$\rightarrow$de and de$\rightarrow$en), then train on the subsampled dataset. As shown in Figure~\ref{fig:subsampling} (and Table~\ref{tab:subsampling} in Appendix~\ref{appendix:b}), \textbf{\textit{finetuning} on the subsampled dataset only took a small performance hit relative to finetuning on the full dataset, but \textit{pretraining} took a large performance hit}. Thus, it is likely that pretraining performance would continue to improve had we generated a larger \texttt{NewsPaLM} dataset, while we would be unlikely to observe substantial incremental improvements in finetuning performance by increasing the dataset size. Also note that the stability in finetuning performance under subsampling held up despite using the most efficient subset selection method (random, as opposed to e.g., QE filtering), another indicator supporting the high quality of our \texttt{NewsPaLM} dataset.
    
\begin{figure}
    \centering
    \includegraphics[width=\columnwidth]{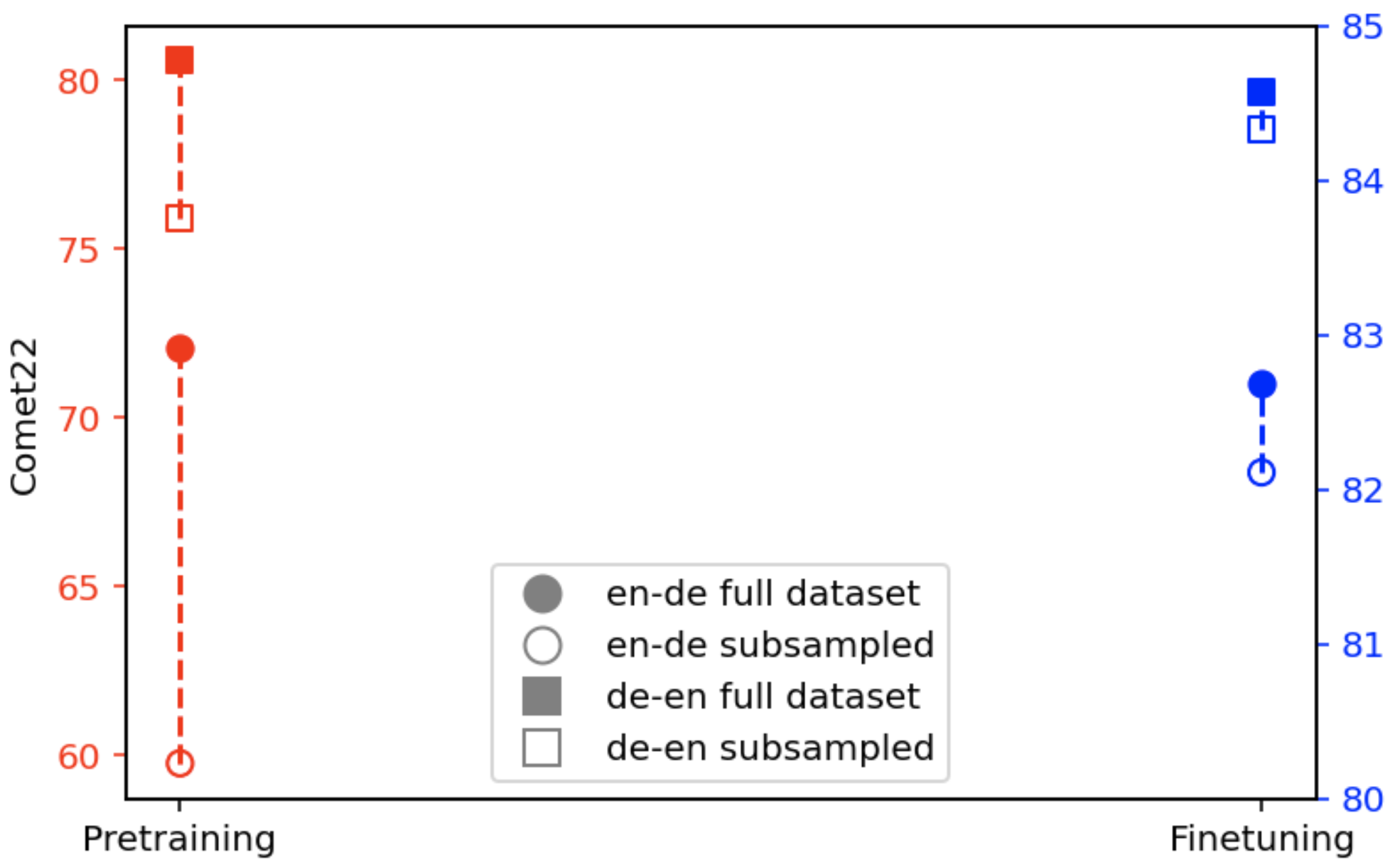}
    \caption{Comparison of model performance when pretraining and finetuning on the full versus subsampled \texttt{NewsPaLM} MBR dataset (WMT'23 test set). The subsampled dataset is 25\% of the size of the full dataset, and was sampled randomly. Note that pretraining performance drops substantially when training on the subsampled dataset (for both en$\rightarrow$de and de$\rightarrow$en), while finetuning performance is minimally affected.}
    \label{fig:subsampling}
\end{figure}

\subsubsection{Effect of Cluster-based Data Selection}
As described in \S\ref{sec:maat}, we used a clustering-based approach (sampling uniformly over the computed clusters) to select the subset of Newscrawl data which we used to generate the \texttt{NewsPaLM} dataset. To isolate the effect of our sample selection technique, we compare its performance against sampling uniformly from the original Newcrawl dataset distribution (i.e., without taking cluster information into account). Since our \texttt{NewsPaLM} dataset contains the subset of Newscrawl examples selected by sampling uniformly over the clusters, we approximate the above comparison as follows, selecting 25\% of the \texttt{NewsPaLM} dataset in both cases:
\begin{itemize}
    \item We sample uniformly from \texttt{NewsPaLM} to approximate cluster-guided sampling from the full Newscrawl dataset.
    \item We use the original cluster sizes of the full Newscrawl dataset (computed prior to selecting the \texttt{NewsPaLM} subset), and sample from \texttt{NewsPaLM} according to this distribution. This approximates sampling from the full Newscrawl dataset \textit{without} taking cluster information into account. Note that because the original cluster distribution was highly skewed, with most of the examples belonging to the top few clusters, we could not exactly match the original distribution while sampling 25\% of the \texttt{NewsPaLM} dataset, but we chose the distribution to be the one which was closest to the original.
\end{itemize}

As shown in Table \ref{tab:cluster-ablation}, using the cluster information in the subsampling procedure marginally improves results for en$\rightarrow$de pretraining and finetuning, and for de$\rightarrow$ en finetuning. (There is no clear signal for de$\rightarrow$en pretraining; according to \textit{MetricX}, using the cluster information helps, while according to \textit{Comet22}, it hurts.)
\begin{table*}[t]
    \centering
    \begin{adjustbox}{width=\textwidth}
\begin{tabular}{llcc}
\toprule
& \bf Model & \bf MetricX $\downarrow$ & \bf COMET22 $\uparrow$ \\
\midrule
\multirow{4}{*}{\textit{en$\rightarrow$de}}
& 1a) PT: Sampling uniformly over clusters & \bf 11.02 & \bf 59.75 \\
& 1b) PT: Sampling uniformly from original Newscrawl distribution & 11.60 & 59.09 \\
\cmidrule{2-4}
& 2a) FT: Sampling uniformly over clusters & \bf 2.55 & \bf 82.11 \\
& 2b) FT:  Sampling uniformly from original Newscrawl distribution & \bf 2.55 & 81.88 \\
\midrule[1.5 pt]
\multirow{4}{*}{\textit{de$\rightarrow$en}}
& 1a) PT: Sampling uniformly over clusters & \bf 7.41 & 75.91 \\
& 1b) PT: Sampling uniformly from original Newscrawl distribution & 7.58 & \bf 76.55 \\
\cmidrule{2-4}
& 2a) FT: Sampling uniformly over clusters & \bf 2.96 & \bf 84.33 \\
& 2b) FT:  Sampling uniformly from original Newscrawl distribution & 3.09 & 83.92 \\
\bottomrule
\end{tabular}
\end{adjustbox}
\caption{Comparison of model performance when trained on subsampled \texttt{NewsPaLM} data with and without cluster-based data selection (WMT'23 test set). PT stands for Pretraining and FT, for Finetuning. Random subsampling of \texttt{NewsPaLM} approximates sampling uniformly across the Newscrawl clusters, while subsampling according to the Newscrawl cluster distribution approximates discarding cluster information and sampling randomly according to the original data distribution.}
\label{tab:cluster-ablation}
\end{table*}
\section{Discussion}
\label{sec:discussion}
Training on LLM-generated, MBR-decoded and QE-reranked datasets is an established technique for leveraging monolingual data to improve NMT model quality (\citet{finkelstein2023mbr}, \citet{wang2024don}). While this technique is highly effective, generating such datasets remains a substantial bottleneck, and is often prohibitively expensive. This work accompanies the first-ever open-source release of a LLM-generated, sentence-level and blob-level MBR and QE dataset. We measure the quality of our dataset in terms of its downstream impact on NMT model performance, both when training a NMT model from scratch and when finetuning.

We find that training from scratch on our MBR-decoded and QE-reranked \texttt{NewsPaLM} dataset outperforms training on the entire WMT'23 training dataset (which is 300 times larger), and also outperforms training on the top-quality subset of the WMT'23 training data (selected via QE filtering, and matching the size of our dataset). Moreover, we find that NMT models are unable to generalize well to multi-sentence queries without exposure to such data at training time, motivating the inclusion of blob-level data in our dataset.

We also find that MBR and QE finetuning outperform finetuning on the greedy-decoded version of our dataset. Unlike \citet{finkelstein2023mbr}, which only performs MBR and QE self-distillation using a small encoder-decoder NMT model, here we show that self-MBR and self-QE finetuning are effective for the much stronger \textit{PaLM-2 Bison} LLM as well.

Finally, we show via subsampling experiments on our \texttt{NewsPaLM} dataset that pretraining versus finetuning performance scale very differently with dataset size: While finetuning performance only took a small hit when reducing our dataset to 25\% of its original size, pretraining performance took a large hit. However, note that the full \texttt{NewsPaLM} dataset is already orders of magnitude smaller than most datasets used for NMT model training, including the WMT'23 training dataset.

\section{Related Work}

While MT research has traditionally relied on MAP decoding or generating $k$-best lists through beam search for MBR decoding, \citet{eikema-aziz-2020-map} proposed an approximation of MBR decoding via unbiased sampling.
Their method aims to address the limitations of MAP decoding \citep{eikema-aziz-2020-map, muller-sennrich-2021-understanding, eikema2022samplingbased} by demonstrating that samples drawn from the NMT model align more faithfully with training data statistics when compared to beam search.
\citet{freitag2022high} showed that using neural metrics results in significant improvements in translation quality.
To the best of our knowledge, this is the first work that applies MBR decoding beyond the sentence level for the task of machine translation.

While the improvements in translation quality afforded by MBR are widely acknowledged, its high computational cost limits its application in practice.
Different approaches have been proposed to speed up MBR computation, e.g.\ \cite{eikema2022samplingbased, cheng2023faster, jinnai2024hyperparameterfree, vamvas2024lineartime, tomani2024qualityaware}.
Similar in spirit to MBR decoding, QE-rescoring approaches~\citep{fernandes-etal-2022-quality} also directly optimize a utility function, with linear-time cost.

We approach the efficiency problem from a different perspective, carrying out a one-off expensive MBR decoding run, which can then be re-used for training and finetuning other models via knowledge distillation \cite{bucilua2006model,hinton2015distilling}.
This technique has been a successful way to improve smaller systems by leveraging the capacities of bigger models, while retaining higher computational efficiency.
The technique has been applied to numerous NLP tasks, including neural machine translation \cite[inter alia]{kim-rush-2016-sequence,tan2018multilingual,zhang2019future,jooste2022knowledge,WAN2024101583}.
In the current era of LLMs, these models provide prime candidates for leveraging their impressive capabilities to improve other models.
\citet{yoo-etal-2021-gpt3mix-leveraging} use GPT3 for data augmentation in different classification tasks, in addition to using soft-labels predicted by the language model.
\citet{hsieh-etal-2023-distilling} propose to use ``rationales'' generated by PaLM to train a much smaller T5 model, achieving comparable or even superior performance.
\citet{li-etal-2024-mt} use "selective" distillation to generate synthetic data using a variant of LLaMA-7B, expanding the coverage of training data for a translation model.
Closer to our work, \citet{finkelstein2023mbr} propose to use MBR on a LLM to generate high-quality translations with which to train a dedicated translation model.
As reference, \cite{xu2024survey} provides a much more comprehensive survey of knowledge distillation approaches using LLMs.

Another dimension related to our work is the area of data selection for NMT training.
While a big amount of work has been focused on filtering noise from web-crawled data \cite[e.g.\ ][]{zaragoza-bernabeu-etal-2022-bicleaner}, there are also approaches aimed at improving the translation quality by limiting the training data to high-quality samples.
\citet{carpuat-etal-2017-detecting} use semantic divergence to select the most relevant portion of the training data, while \cite{peter2023there} use QE metrics on the training data to select only high-quality sentence pairs.
\citet{xu2023paradigm} indeed show that only a small amount of high-quality multilingual and parallel data is needed for obtaining state-of-the art translation results finetuning a LLM.
A similar approach was used by \cite{alves2024tower} to finetune LLaMA for translation and translation-related tasks.

One can also find different examples of clustering for data selection for NLP tasks.
\citet{aharoni-goldberg-2020-unsupervised} showed that automatic clustering techniques can adequately recover semantic information from text corpora.
\citet{yu-etal-2023-cold} use clustering for data selection to finetuning a LLM.
Related to these approaches, nearest-neighbor machine translation \cite{khandelwalnearest} uses distance measures between examples to select examples closer to the sentence to translate in an additional module of a translation system.
\cite{agrawal-etal-2023-context} and \cite{vilar-etal-2023-prompting} use similar approaches to construct prompts for LLMs.
\section{Conclusion}

In this work, we have described the dataset creation process for the first-ever release of a LLM-generated MBR and QE dataset.
We have shown that this dataset can be used to build a small and efficient, but high-quality, NMT model from scratch.
In fact, training on this dataset outperforms training on the much larger, human-generated WMT'23 dataset.
We have also shown that this dataset can improve NMT performance during finetuning, both for an encoder-decoder system and via self-distillation for an already highly performant LLM.
We hope that this dataset will enable further distillation research by the wider community, even by those without resources to generate datasets from large teacher models using expensive decoding techniques. 

There are many avenues for future work.
This work presented the first investigation of multi-sentence (i.e., blob-level) QE finetuning, and a natural next step would be to move to the document level.
The dataset creation process described here could also be continued iteratively, by generating a new MBR and QE dataset from the same LLM teacher, but after finetuning on the original version of the dataset (or the version from the previous iteration).
While this would be expensive, it would likely yield further incremental improvements in dataset quality.
Finally, there remain many open questions regarding how to optimally perform distillation of a stronger teacher model into a weaker student using MBR and QE data.
For instance, rather than finetuning on a uniform mixture of all examples in the dataset, the student model's perplexity on these examples could be taken into account to select a subset of examples and/or to determine the optimal progression of examples to expose the student to during finetuning.

\clearpage
\section*{Limitations}

The (target-side) data generation process was expensive, due to both using a LLM and a costly decoding method. For MBR dataset creation, computation of each dataset example required generation of $n$ outputs from the LLM teacher model, and then $O(n^2)$ forward passes through the utility function, where $n$ is the candidate size. For QE dataset generation, $O(n)$ forward passes through the utility function were required per example. Thus, the dataset construction method proposed here is not easily scalable to other language pairs and/or source-side data in the absence of substantial computing resources.

\bibliography{custom}

\begin{thebibliography}{51}
\expandafter\ifx\csname natexlab\endcsname\relax\def\natexlab#1{#1}\fi

\bibitem[{Agrawal et~al.(2023)Agrawal, Zhou, Lewis, Zettlemoyer, and
  Ghazvininejad}]{agrawal-etal-2023-context}
Sweta Agrawal, Chunting Zhou, Mike Lewis, Luke Zettlemoyer, and Marjan
  Ghazvininejad. 2023.
\newblock \href {https://doi.org/10.18653/v1/2023.findings-acl.564} {In-context
  examples selection for machine translation}.
\newblock In \emph{Findings of the Association for Computational Linguistics:
  ACL 2023}, pages 8857--8873, Toronto, Canada. Association for Computational
  Linguistics.

\bibitem[{Aharoni and Goldberg(2020)}]{aharoni-goldberg-2020-unsupervised}
Roee Aharoni and Yoav Goldberg. 2020.
\newblock \href {https://doi.org/10.18653/v1/2020.acl-main.692} {Unsupervised
  domain clusters in pretrained language models}.
\newblock In \emph{Proceedings of the 58th Annual Meeting of the Association
  for Computational Linguistics}, pages 7747--7763, Online. Association for
  Computational Linguistics.

\bibitem[{Akhbardeh et~al.(2021)Akhbardeh, Arkhangorodsky, Biesialska, Bojar,
  Chatterjee, Chaudhary, Costa-jussa, Espa{\~n}a-Bonet, Fan, Federmann,
  Freitag, Graham, Grundkiewicz, Haddow, Harter, Heafield, Homan, Huck,
  Amponsah-Kaakyire, Kasai, Khashabi, Knight, Kocmi, Koehn, Lourie, Monz,
  Morishita, Nagata, Nagesh, Nakazawa, Negri, Pal, Tapo, Turchi, Vydrin, and
  Zampieri}]{akhbardeh-etal-2021-findings}
Farhad Akhbardeh, Arkady Arkhangorodsky, Magdalena Biesialska, Ond{\v{r}}ej
  Bojar, Rajen Chatterjee, Vishrav Chaudhary, Marta~R. Costa-jussa, Cristina
  Espa{\~n}a-Bonet, Angela Fan, Christian Federmann, Markus Freitag, Yvette
  Graham, Roman Grundkiewicz, Barry Haddow, Leonie Harter, Kenneth Heafield,
  Christopher Homan, Matthias Huck, Kwabena Amponsah-Kaakyire, Jungo Kasai,
  Daniel Khashabi, Kevin Knight, Tom Kocmi, Philipp Koehn, Nicholas Lourie,
  Christof Monz, Makoto Morishita, Masaaki Nagata, Ajay Nagesh, Toshiaki
  Nakazawa, Matteo Negri, Santanu Pal, Allahsera~Auguste Tapo, Marco Turchi,
  Valentin Vydrin, and Marcos Zampieri. 2021.
\newblock \href {https://aclanthology.org/2021.wmt-1.1} {Findings of the 2021
  conference on machine translation ({WMT}21)}.
\newblock In \emph{Proceedings of the Sixth Conference on Machine Translation},
  pages 1--88, Online. Association for Computational Linguistics.

\bibitem[{Alves et~al.(2024)Alves, Pombal, Guerreiro, Martins, Alves, Farajian,
  Peters, Rei, Fernandes, Agrawal et~al.}]{alves2024tower}
Duarte~M Alves, Jos{\'e} Pombal, Nuno~M Guerreiro, Pedro~H Martins, Jo{\~a}o
  Alves, Amin Farajian, Ben Peters, Ricardo Rei, Patrick Fernandes, Sweta
  Agrawal, et~al. 2024.
\newblock Tower: An open multilingual large language model for
  translation-related tasks.
\newblock \emph{arXiv preprint arXiv:2402.17733}.

\bibitem[{Anil et~al.(2023)Anil, Dai, Firat, Johnson, Lepikhin, Passos,
  Shakeri, Taropa, Bailey, Chen, Chu, Clark, Shafey, Huang, Meier-Hellstern,
  Mishra, Moreira, Omernick, Robinson, Ruder, Tay, Xiao, Xu, Zhang, Abrego,
  Ahn, Austin, Barham, Botha, Bradbury, Brahma, Brooks, Catasta, Cheng, Cherry,
  Choquette-Choo, Chowdhery, Crepy, Dave, Dehghani, Dev, Devlin, Díaz, Du,
  Dyer, Feinberg, Feng, Fienber, Freitag, Garcia, Gehrmann, Gonzalez, Gur-Ari,
  Hand, Hashemi, Hou, Howland, Hu, Hui, Hurwitz, Isard, Ittycheriah, Jagielski,
  Jia, Kenealy, Krikun, Kudugunta, Lan, Lee, Lee, Li, Li, Li, Li, Li, Lim, Lin,
  Liu, Liu, Maggioni, Mahendru, Maynez, Misra, Moussalem, Nado, Nham, Ni,
  Nystrom, Parrish, Pellat, Polacek, Polozov, Pope, Qiao, Reif, Richter, Riley,
  Ros, Roy, Saeta, Samuel, Shelby, Slone, Smilkov, So, Sohn, Tokumine, Valter,
  Vasudevan, Vodrahalli, Wang, Wang, Wang, Wang, Wieting, Wu, Xu, Xu, Xue, Yin,
  Yu, Zhang, Zheng, Zheng, Zhou, Zhou, Petrov, and Wu}]{anil2023palm}
Rohan Anil, Andrew~M. Dai, Orhan Firat, Melvin Johnson, Dmitry Lepikhin,
  Alexandre Passos, Siamak Shakeri, Emanuel Taropa, Paige Bailey, Zhifeng Chen,
  Eric Chu, Jonathan~H. Clark, Laurent~El Shafey, Yanping Huang, Kathy
  Meier-Hellstern, Gaurav Mishra, Erica Moreira, Mark Omernick, Kevin Robinson,
  Sebastian Ruder, Yi~Tay, Kefan Xiao, Yuanzhong Xu, Yujing Zhang,
  Gustavo~Hernandez Abrego, Junwhan Ahn, Jacob Austin, Paul Barham, Jan Botha,
  James Bradbury, Siddhartha Brahma, Kevin Brooks, Michele Catasta, Yong Cheng,
  Colin Cherry, Christopher~A. Choquette-Choo, Aakanksha Chowdhery, Clément
  Crepy, Shachi Dave, Mostafa Dehghani, Sunipa Dev, Jacob Devlin, Mark Díaz,
  Nan Du, Ethan Dyer, Vlad Feinberg, Fangxiaoyu Feng, Vlad Fienber, Markus
  Freitag, Xavier Garcia, Sebastian Gehrmann, Lucas Gonzalez, Guy Gur-Ari,
  Steven Hand, Hadi Hashemi, Le~Hou, Joshua Howland, Andrea Hu, Jeffrey Hui,
  Jeremy Hurwitz, Michael Isard, Abe Ittycheriah, Matthew Jagielski, Wenhao
  Jia, Kathleen Kenealy, Maxim Krikun, Sneha Kudugunta, Chang Lan, Katherine
  Lee, Benjamin Lee, Eric Li, Music Li, Wei Li, YaGuang Li, Jian Li, Hyeontaek
  Lim, Hanzhao Lin, Zhongtao Liu, Frederick Liu, Marcello Maggioni, Aroma
  Mahendru, Joshua Maynez, Vedant Misra, Maysam Moussalem, Zachary Nado, John
  Nham, Eric Ni, Andrew Nystrom, Alicia Parrish, Marie Pellat, Martin Polacek,
  Alex Polozov, Reiner Pope, Siyuan Qiao, Emily Reif, Bryan Richter, Parker
  Riley, Alex~Castro Ros, Aurko Roy, Brennan Saeta, Rajkumar Samuel, Renee
  Shelby, Ambrose Slone, Daniel Smilkov, David~R. So, Daniel Sohn, Simon
  Tokumine, Dasha Valter, Vijay Vasudevan, Kiran Vodrahalli, Xuezhi Wang,
  Pidong Wang, Zirui Wang, Tao Wang, John Wieting, Yuhuai Wu, Kelvin Xu, Yunhan
  Xu, Linting Xue, Pengcheng Yin, Jiahui Yu, Qiao Zhang, Steven Zheng,
  Ce~Zheng, Weikang Zhou, Denny Zhou, Slav Petrov, and Yonghui Wu. 2023.
\newblock \href {http://arxiv.org/abs/2305.10403} {Palm 2 technical report}.

\bibitem[{Buciluǎ et~al.(2006)Buciluǎ, Caruana, and
  Niculescu-Mizil}]{bucilua2006model}
Cristian Buciluǎ, Rich Caruana, and Alexandru Niculescu-Mizil. 2006.
\newblock Model compression.
\newblock In \emph{Proceedings of the 12th ACM SIGKDD international conference
  on Knowledge discovery and data mining}, pages 535--541.

\bibitem[{Carpuat et~al.(2017)Carpuat, Vyas, and
  Niu}]{carpuat-etal-2017-detecting}
Marine Carpuat, Yogarshi Vyas, and Xing Niu. 2017.
\newblock \href {https://doi.org/10.18653/v1/W17-3209} {Detecting cross-lingual
  semantic divergence for neural machine translation}.
\newblock In \emph{Proceedings of the First Workshop on Neural Machine
  Translation}, pages 69--79, Vancouver. Association for Computational
  Linguistics.

\bibitem[{Cheng and Vlachos(2023)}]{cheng2023faster}
Julius Cheng and Andreas Vlachos. 2023.
\newblock \href {http://arxiv.org/abs/2311.14919} {Faster minimum bayes risk
  decoding with confidence-based pruning}.

\bibitem[{Conneau et~al.(2020)Conneau, Khandelwal, Goyal, Chaudhary, Wenzek,
  Guzm{\'a}n, Grave, Ott, Zettlemoyer, and
  Stoyanov}]{conneau-etal-2020-unsupervised}
Alexis Conneau, Kartikay Khandelwal, Naman Goyal, Vishrav Chaudhary, Guillaume
  Wenzek, Francisco Guzm{\'a}n, Edouard Grave, Myle Ott, Luke Zettlemoyer, and
  Veselin Stoyanov. 2020.
\newblock \href {https://doi.org/10.18653/v1/2020.acl-main.747} {Unsupervised
  cross-lingual representation learning at scale}.
\newblock In \emph{Proceedings of the 58th Annual Meeting of the Association
  for Computational Linguistics}, pages 8440--8451, Online. Association for
  Computational Linguistics.

\bibitem[{Eikema and Aziz(2020{\natexlab{a}})}]{eikema2020map}
Bryan Eikema and Wilker Aziz. 2020{\natexlab{a}}.
\newblock Is map decoding all you need? the inadequacy of the mode in neural
  machine translation.
\newblock \emph{arXiv preprint arXiv:2005.10283}.

\bibitem[{Eikema and Aziz(2020{\natexlab{b}})}]{eikema-aziz-2020-map}
Bryan Eikema and Wilker Aziz. 2020{\natexlab{b}}.
\newblock \href {https://doi.org/10.18653/v1/2020.coling-main.398} {Is {MAP}
  decoding all you need? the inadequacy of the mode in neural machine
  translation}.
\newblock In \emph{Proceedings of the 28th International Conference on
  Computational Linguistics}, pages 4506--4520, Barcelona, Spain (Online).
  International Committee on Computational Linguistics.

\bibitem[{Eikema and Aziz(2022)}]{eikema2022samplingbased}
Bryan Eikema and Wilker Aziz. 2022.
\newblock \href {https://doi.org/10.18653/v1/2022.emnlp-main.754}
  {Sampling-based approximations to minimum {B}ayes risk decoding for neural
  machine translation}.
\newblock In \emph{Proceedings of the 2022 Conference on Empirical Methods in
  Natural Language Processing}, pages 10978--10993, Abu Dhabi, United Arab
  Emirates. Association for Computational Linguistics.

\bibitem[{Farhad et~al.(2021)Farhad, Arkady, Magdalena, Ond{\v{r}}ej, Rajen,
  Vishrav, Costa-jussa, Cristina, Angela, Christian
  et~al.}]{farhad2021findings}
Akhbardeh Farhad, Arkhangorodsky Arkady, Biesialska Magdalena, Bojar
  Ond{\v{r}}ej, Chatterjee Rajen, Chaudhary Vishrav, Marta~R Costa-jussa,
  Espa{\~n}a-Bonet Cristina, Fan Angela, Federmann Christian, et~al. 2021.
\newblock Findings of the 2021 conference on machine translation (wmt21).
\newblock In \emph{Proceedings of the Sixth Conference on Machine Translation},
  pages 1--88. Association for Computational Linguistics.

\bibitem[{Fernandes et~al.(2022)Fernandes, Farinhas, Rei, C.~de Souza, Ogayo,
  Neubig, and Martins}]{fernandes-etal-2022-quality}
Patrick Fernandes, Ant{\'o}nio Farinhas, Ricardo Rei, Jos{\'e}~G. C.~de Souza,
  Perez Ogayo, Graham Neubig, and Andre Martins. 2022.
\newblock \href {https://doi.org/10.18653/v1/2022.naacl-main.100}
  {Quality-aware decoding for neural machine translation}.
\newblock In \emph{Proceedings of the 2022 Conference of the North American
  Chapter of the Association for Computational Linguistics: Human Language
  Technologies}, pages 1396--1412, Seattle, United States. Association for
  Computational Linguistics.

\bibitem[{Finkelstein and Freitag(2023)}]{finkelstein2023mbr}
Mara Finkelstein and Markus Freitag. 2023.
\newblock Mbr and qe finetuning: Training-time distillation of the best and
  most expensive decoding methods.
\newblock \emph{arXiv preprint arXiv:2309.10966}.

\bibitem[{Freitag et~al.(2023)Freitag, Ghorbani, and
  Fernandes}]{freitag2023epsilon}
Markus Freitag, Behrooz Ghorbani, and Patrick Fernandes. 2023.
\newblock Epsilon sampling rocks: Investigating sampling strategies for minimum
  bayes risk decoding for machine translation.
\newblock \emph{arXiv preprint arXiv:2305.09860}.

\bibitem[{Freitag et~al.(2022)Freitag, Grangier, Tan, and
  Liang}]{freitag2022high}
Markus Freitag, David Grangier, Qijun Tan, and Bowen Liang. 2022.
\newblock High quality rather than high model probability: Minimum bayes risk
  decoding with neural metrics.
\newblock \emph{Transactions of the Association for Computational Linguistics},
  10:811--825.

\bibitem[{{Gemini Team}(2024)}]{geminiteam2024geminifamilyhighlycapable}
{Gemini Team}. 2024.
\newblock \href {http://arxiv.org/abs/2312.11805} {Gemini: A family of highly
  capable multimodal models}.

\bibitem[{Hewitt et~al.(2022)Hewitt, Manning, and Liang}]{hewitt2022truncation}
John Hewitt, Christopher~D Manning, and Percy Liang. 2022.
\newblock Truncation sampling as language model desmoothing.
\newblock \emph{arXiv preprint arXiv:2210.15191}.

\bibitem[{Hinton et~al.(2015)Hinton, Vinyals, and Dean}]{hinton2015distilling}
Geoffrey Hinton, Oriol Vinyals, and Jeff Dean. 2015.
\newblock Distilling the knowledge in a neural network.
\newblock \emph{arXiv preprint arXiv:1503.02531}.

\bibitem[{Hsieh et~al.(2023)Hsieh, Li, Yeh, Nakhost, Fujii, Ratner, Krishna,
  Lee, and Pfister}]{hsieh-etal-2023-distilling}
Cheng-Yu Hsieh, Chun-Liang Li, Chih-kuan Yeh, Hootan Nakhost, Yasuhisa Fujii,
  Alex Ratner, Ranjay Krishna, Chen-Yu Lee, and Tomas Pfister. 2023.
\newblock \href {https://doi.org/10.18653/v1/2023.findings-acl.507} {Distilling
  step-by-step! outperforming larger language models with less training data
  and smaller model sizes}.
\newblock In \emph{Findings of the Association for Computational Linguistics:
  ACL 2023}, pages 8003--8017, Toronto, Canada. Association for Computational
  Linguistics.

\bibitem[{Jinnai and Ariu(2024)}]{jinnai2024hyperparameterfree}
Yuu Jinnai and Kaito Ariu. 2024.
\newblock \href {http://arxiv.org/abs/2401.02749} {Hyperparameter-free approach
  for faster minimum bayes risk decoding}.

\bibitem[{Jooste et~al.(2022)Jooste, Haque, and Way}]{jooste2022knowledge}
Wandri Jooste, Rejwanul Haque, and Andy Way. 2022.
\newblock Knowledge distillation: A method for making neural machine
  translation more efficient.
\newblock \emph{Information}, 13(2):88.

\bibitem[{Juraska et~al.(2023)Juraska, Finkelstein, Deutsch, Siddhant,
  Mirzazadeh, and Freitag}]{juraska-etal-2023-metricx}
Juraj Juraska, Mara Finkelstein, Daniel Deutsch, Aditya Siddhant, Mehdi
  Mirzazadeh, and Markus Freitag. 2023.
\newblock \href {https://doi.org/10.18653/v1/2023.wmt-1.63} {{MetricX-23: The
  Google Submission to the WMT 2023 Metrics Shared Task}}.
\newblock In \emph{Proceedings of the Eighth Conference on Machine
  Translation}, pages 756--767, Singapore. Association for Computational
  Linguistics.

\bibitem[{Khandelwal et~al.(2021)Khandelwal, Fan, Jurafsky, Zettlemoyer, and
  Lewis}]{khandelwalnearest}
Urvashi Khandelwal, Angela Fan, Dan Jurafsky, Luke Zettlemoyer, and Mike Lewis.
  2021.
\newblock Nearest neighbor machine translation.
\newblock In \emph{International Conference on Learning Representations}.

\bibitem[{Kim and Rush(2016)}]{kim-rush-2016-sequence}
Yoon Kim and Alexander~M. Rush. 2016.
\newblock \href {https://doi.org/10.18653/v1/D16-1139} {Sequence-level
  knowledge distillation}.
\newblock In \emph{Proceedings of the 2016 Conference on Empirical Methods in
  Natural Language Processing}, pages 1317--1327, Austin, Texas. Association
  for Computational Linguistics.

\bibitem[{Kocmi et~al.(2024{\natexlab{a}})Kocmi, Avramidis, Bawden, Bojar,
  Dvorkovich, Federmann, Fishel, Freitag, Gowda, Grundkiewicz, Haddow,
  Karpinska, Koehn, Marie, Murray, Nagata, Popel, Popovic, Shmatova,
  Steingrímsson, and Zouhar}]{kocmi2024preliminarywmt24rankinggeneral}
Tom Kocmi, Eleftherios Avramidis, Rachel Bawden, Ondrej Bojar, Anton
  Dvorkovich, Christian Federmann, Mark Fishel, Markus Freitag, Thamme Gowda,
  Roman Grundkiewicz, Barry Haddow, Marzena Karpinska, Philipp Koehn, Benjamin
  Marie, Kenton Murray, Masaaki Nagata, Martin Popel, Maja Popovic, Mariya
  Shmatova, Steinþór Steingrímsson, and Vilém Zouhar. 2024{\natexlab{a}}.
\newblock \href {http://arxiv.org/abs/2407.19884} {Preliminary wmt24 ranking of
  general mt systems and llms}.

\bibitem[{Kocmi et~al.(2023{\natexlab{a}})Kocmi, Avramidis, Bawden, Bojar,
  Dvorkovich, Federmann, Fishel, Freitag, Gowda, Grundkiewicz, Haddow, Koehn,
  Marie, Monz, Morishita, Murray, Nagata, Nakazawa, Popel, Popovi{\'c}, and
  Shmatova}]{kocmi-etal-2023-findings}
Tom Kocmi, Eleftherios Avramidis, Rachel Bawden, Ond{\v{r}}ej Bojar, Anton
  Dvorkovich, Christian Federmann, Mark Fishel, Markus Freitag, Thamme Gowda,
  Roman Grundkiewicz, Barry Haddow, Philipp Koehn, Benjamin Marie, Christof
  Monz, Makoto Morishita, Kenton Murray, Makoto Nagata, Toshiaki Nakazawa,
  Martin Popel, Maja Popovi{\'c}, and Mariya Shmatova. 2023{\natexlab{a}}.
\newblock \href {https://doi.org/10.18653/v1/2023.wmt-1.1} {Findings of the
  2023 conference on machine translation ({WMT}23): {LLM}s are here but not
  quite there yet}.
\newblock In \emph{Proceedings of the Eighth Conference on Machine
  Translation}, pages 1--42, Singapore. Association for Computational
  Linguistics.

\bibitem[{Kocmi et~al.(2023{\natexlab{b}})Kocmi, Avramidis, Bawden, Bojar,
  Dvorkovich, Federmann, Fishel, Freitag, Gowda, Grundkiewicz
  et~al.}]{kocmi2023findings}
Tom Kocmi, Eleftherios Avramidis, Rachel Bawden, Ond{\v{r}}ej Bojar, Anton
  Dvorkovich, Christian Federmann, Mark Fishel, Markus Freitag, Thamme Gowda,
  Roman Grundkiewicz, et~al. 2023{\natexlab{b}}.
\newblock Findings of the 2023 conference on machine translation (wmt23): Llms
  are here but not quite there yet.
\newblock In \emph{Proceedings of the Eighth Conference on Machine
  Translation}, pages 1--42.

\bibitem[{Kocmi et~al.(2024{\natexlab{b}})Kocmi, Avramidis, Bawden, Bojar,
  Dvorkovich, Federmann, Fishel, Freitag, Gowda, Grundkiewicz
  et~al.}]{kocmi2024preliminary}
Tom Kocmi, Eleftherios Avramidis, Rachel Bawden, Ondrej Bojar, Anton
  Dvorkovich, Christian Federmann, Mark Fishel, Markus Freitag, Thamme Gowda,
  Roman Grundkiewicz, et~al. 2024{\natexlab{b}}.
\newblock Preliminary wmt24 ranking of general mt systems and llms.
\newblock \emph{arXiv preprint arXiv:2407.19884}.

\bibitem[{Kocmi et~al.(2022)Kocmi, Bawden, Bojar, Dvorkovich, Federmann,
  Fishel, Gowda, Graham, Grundkiewicz, Haddow, Knowles, Koehn, Monz, Morishita,
  Nagata, Nakazawa, Nov{\'a}k, Popel, and
  Popovi{\'c}}]{kocmi-etal-2022-findings}
Tom Kocmi, Rachel Bawden, Ond{\v{r}}ej Bojar, Anton Dvorkovich, Christian
  Federmann, Mark Fishel, Thamme Gowda, Yvette Graham, Roman Grundkiewicz,
  Barry Haddow, Rebecca Knowles, Philipp Koehn, Christof Monz, Makoto
  Morishita, Masaaki Nagata, Toshiaki Nakazawa, Michal Nov{\'a}k, Martin Popel,
  and Maja Popovi{\'c}. 2022.
\newblock \href {https://aclanthology.org/2022.wmt-1.1} {Findings of the 2022
  conference on machine translation ({WMT}22)}.
\newblock In \emph{Proceedings of the Seventh Conference on Machine Translation
  (WMT)}, pages 1--45, Abu Dhabi, United Arab Emirates (Hybrid). Association
  for Computational Linguistics.

\bibitem[{Li et~al.(2024)Li, Cheng, Huang, and Chen}]{li-etal-2024-mt}
Jiahuan Li, Shanbo Cheng, Shujian Huang, and Jiajun Chen. 2024.
\newblock \href {https://doi.org/10.18653/v1/2024.naacl-long.358}
  {{MT}-{PATCHER}: Selective and extendable knowledge distillation from large
  language models for machine translation}.
\newblock In \emph{Proceedings of the 2024 Conference of the North American
  Chapter of the Association for Computational Linguistics: Human Language
  Technologies (Volume 1: Long Papers)}, pages 6445--6459, Mexico City, Mexico.
  Association for Computational Linguistics.

\bibitem[{M{\"u}ller and Sennrich(2021)}]{muller-sennrich-2021-understanding}
Mathias M{\"u}ller and Rico Sennrich. 2021.
\newblock \href {https://doi.org/10.18653/v1/2021.acl-long.22} {Understanding
  the properties of minimum {B}ayes risk decoding in neural machine
  translation}.
\newblock In \emph{Proceedings of the 59th Annual Meeting of the Association
  for Computational Linguistics and the 11th International Joint Conference on
  Natural Language Processing (Volume 1: Long Papers)}, pages 259--272, Online.
  Association for Computational Linguistics.

\bibitem[{Peter et~al.(2023)Peter, Vilar, Deutsch, Finkelstein, Juraska, and
  Freitag}]{peter2023there}
Jan-Thorsten Peter, David Vilar, Daniel Deutsch, Mara Finkelstein, Juraj
  Juraska, and Markus Freitag. 2023.
\newblock There's no data like better data: Using qe metrics for mt data
  filtering.
\newblock \emph{arXiv preprint arXiv:2311.05350}.

\bibitem[{Rei et~al.(2022)Rei, De~Souza, Alves, Zerva, Farinha, Glushkova,
  Lavie, Coheur, and Martins}]{rei2022comet}
Ricardo Rei, Jos{\'e}~GC De~Souza, Duarte Alves, Chrysoula Zerva, Ana~C
  Farinha, Taisiya Glushkova, Alon Lavie, Luisa Coheur, and Andr{\'e}~FT
  Martins. 2022.
\newblock Comet-22: Unbabel-ist 2022 submission for the metrics shared task.
\newblock In \emph{Proceedings of the Seventh Conference on Machine Translation
  (WMT)}, pages 578--585.

\bibitem[{Rei et~al.(2020)Rei, Stewart, Farinha, and Lavie}]{rei2020comet}
Ricardo Rei, Craig Stewart, Ana~C Farinha, and Alon Lavie. 2020.
\newblock Comet: A neural framework for mt evaluation.
\newblock \emph{arXiv preprint arXiv:2009.09025}.

\bibitem[{Sellam et~al.(2020)Sellam, Das, and Parikh}]{sellam2020bleurt}
Thibault Sellam, Dipanjan Das, and Ankur~P Parikh. 2020.
\newblock Bleurt: Learning robust metrics for text generation.
\newblock \emph{arXiv preprint arXiv:2004.04696}.

\bibitem[{Sumengen et~al.(2021)Sumengen, Rajagopalan, Citovsky, Simcha, Bachem,
  Mitra, Blasiak, Liang, and Kumar}]{sumengen2021scaling}
Baris Sumengen, Anand Rajagopalan, Gui Citovsky, David Simcha, Olivier Bachem,
  Pradipta Mitra, Sam Blasiak, Mason Liang, and Sanjiv Kumar. 2021.
\newblock Scaling hierarchical agglomerative clustering to billion-sized
  datasets.
\newblock \emph{arXiv preprint arXiv:2105.11653}.

\bibitem[{Tan et~al.(2018)Tan, Ren, He, Qin, Zhao, and
  Liu}]{tan2018multilingual}
Xu~Tan, Yi~Ren, Di~He, Tao Qin, Zhou Zhao, and Tie-Yan Liu. 2018.
\newblock Multilingual neural machine translation with knowledge distillation.
\newblock In \emph{International Conference on Learning Representations}.

\bibitem[{Tomani et~al.(2024)Tomani, Vilar, Freitag, Cherry, Naskar,
  Finkelstein, Garcia, and Cremers}]{tomani2024qualityaware}
Christian Tomani, David Vilar, Markus Freitag, Colin Cherry, Subhajit Naskar,
  Mara Finkelstein, Xavier Garcia, and Daniel Cremers. 2024.
\newblock \href {http://arxiv.org/abs/2310.06707} {Quality-aware translation
  models: Efficient generation and quality estimation in a single model}.

\bibitem[{Vamvas and Sennrich(2024)}]{vamvas2024lineartime}
Jannis Vamvas and Rico Sennrich. 2024.
\newblock \href {http://arxiv.org/abs/2402.04251} {Linear-time minimum bayes
  risk decoding with reference aggregation}.

\bibitem[{Vaswani et~al.(2017)Vaswani, Shazeer, Parmar, Uszkoreit, Jones,
  Gomez, Kaiser, and Polosukhin}]{vaswani2017attention}
Ashish Vaswani, Noam Shazeer, Niki Parmar, Jakob Uszkoreit, Llion Jones,
  Aidan~N Gomez, {\L}ukasz Kaiser, and Illia Polosukhin. 2017.
\newblock Attention is all you need.
\newblock \emph{Advances in neural information processing systems}, 30.

\bibitem[{Vilar et~al.(2023)Vilar, Freitag, Cherry, Luo, Ratnakar, and
  Foster}]{vilar-etal-2023-prompting}
David Vilar, Markus Freitag, Colin Cherry, Jiaming Luo, Viresh Ratnakar, and
  George Foster. 2023.
\newblock \href {https://doi.org/10.18653/v1/2023.acl-long.859} {Prompting
  {P}a{LM} for translation: Assessing strategies and performance}.
\newblock In \emph{Proceedings of the 61st Annual Meeting of the Association
  for Computational Linguistics (Volume 1: Long Papers)}, pages 15406--15427,
  Toronto, Canada. Association for Computational Linguistics.

\bibitem[{Wan et~al.(2024)Wan, Zhang, Li, Zhang, and Li}]{WAN2024101583}
Yuxian Wan, Wenlin Zhang, Zhen Li, Hao Zhang, and Yanxia Li. 2024.
\newblock \href {https://doi.org/https://doi.org/10.1016/j.csl.2023.101583}
  {Dual knowledge distillation for neural machine translation}.
\newblock \emph{Computer Speech \& Language}, 84:101583.

\bibitem[{Wang et~al.(2024)Wang, Briakou, Dadkhahi, Agarwal, Cherry, and
  Cohn}]{wang2024don}
Jun Wang, Eleftheria Briakou, Hamid Dadkhahi, Rishabh Agarwal, Colin Cherry,
  and Trevor Cohn. 2024.
\newblock Don't throw away data: Better sequence knowledge distillation.
\newblock \emph{arXiv preprint arXiv:2407.10456}.

\bibitem[{Xu et~al.(2023)Xu, Kim, Sharaf, and Awadalla}]{xu2023paradigm}
Haoran Xu, Young~Jin Kim, Amr Sharaf, and Hany~Hassan Awadalla. 2023.
\newblock A paradigm shift in machine translation: Boosting translation
  performance of large language models.
\newblock \emph{arXiv preprint arXiv:2309.11674}.

\bibitem[{Xu et~al.(2024)Xu, Li, Tao, Shen, Cheng, Li, Xu, Tao, and
  Zhou}]{xu2024survey}
Xiaohan Xu, Ming Li, Chongyang Tao, Tao Shen, Reynold Cheng, Jinyang Li, Can
  Xu, Dacheng Tao, and Tianyi Zhou. 2024.
\newblock A survey on knowledge distillation of large language models.
\newblock \emph{arXiv preprint arXiv:2402.13116}.

\bibitem[{Yoo et~al.(2021)Yoo, Park, Kang, Lee, and
  Park}]{yoo-etal-2021-gpt3mix-leveraging}
Kang~Min Yoo, Dongju Park, Jaewook Kang, Sang-Woo Lee, and Woomyoung Park.
  2021.
\newblock \href {https://doi.org/10.18653/v1/2021.findings-emnlp.192}
  {{GPT}3{M}ix: Leveraging large-scale language models for text augmentation}.
\newblock In \emph{Findings of the Association for Computational Linguistics:
  EMNLP 2021}, pages 2225--2239, Punta Cana, Dominican Republic. Association
  for Computational Linguistics.

\bibitem[{Yu et~al.(2023)Yu, Zhang, Xu, Zhang, Shen, and
  Zhang}]{yu-etal-2023-cold}
Yue Yu, Rongzhi Zhang, Ran Xu, Jieyu Zhang, Jiaming Shen, and Chao Zhang. 2023.
\newblock \href {https://doi.org/10.18653/v1/2023.acl-long.141} {Cold-start
  data selection for better few-shot language model fine-tuning: A prompt-based
  uncertainty propagation approach}.
\newblock In \emph{Proceedings of the 61st Annual Meeting of the Association
  for Computational Linguistics (Volume 1: Long Papers)}, pages 2499--2521,
  Toronto, Canada. Association for Computational Linguistics.

\bibitem[{Zaragoza-Bernabeu et~al.(2022)Zaragoza-Bernabeu,
  Ram{\'\i}rez-S{\'a}nchez, Ba{\~n}{\'o}n, and
  Ortiz~Rojas}]{zaragoza-bernabeu-etal-2022-bicleaner}
Jaume Zaragoza-Bernabeu, Gema Ram{\'\i}rez-S{\'a}nchez, Marta Ba{\~n}{\'o}n,
  and Sergio Ortiz~Rojas. 2022.
\newblock \href {https://aclanthology.org/2022.lrec-1.87} {Bicleaner {AI}:
  Bicleaner goes neural}.
\newblock In \emph{Proceedings of the Thirteenth Language Resources and
  Evaluation Conference}, pages 824--831, Marseille, France. European Language
  Resources Association.

\bibitem[{Zhang et~al.(2019)Zhang, Xiong, Su, and Luo}]{zhang2019future}
Biao Zhang, Deyi Xiong, Jinsong Su, and Jiebo Luo. 2019.
\newblock Future-aware knowledge distillation for neural machine translation.
\newblock \emph{IEEE/ACM Transactions on Audio, Speech, and Language
  Processing}, 27(12):2278--2287.

\end{thebibliography}
\bibliographystyle{acl_natbib}

\clearpage
\appendix
\section{Additional Dataset Statistics}
\label{appendix:a}

Table~\ref{tab:length_ratios} shows the average source-to-target length ratios for each of the \texttt{NewsPaLM} MBR and QE datasets. (See Table~\ref{tab:src_tgt_lengths} in Section~\ref{sec:datasets} for the average source lengths and average targets lengths of the \texttt{NewsPaLM} datasets.)

\begin{table}[!htb]
    \centering
    \begin{adjustbox}{width=\columnwidth}
\begin{tabular}{ccc}
    \toprule
     & \bf Sentence-level & \bf Blob-level \\
    \midrule
    \textsc{EN $\rightarrow$ DE} & 0.9604 & 1.0812 \\
    \textsc{DE $\rightarrow$ EN} & 0.9009 & 0.9729 \\
    \bottomrule
\end{tabular}
    \end{adjustbox}
    \caption{Source-to-target length ratios per dataset, computed using the \texttt{Moses} tokenizer.}
    \label{tab:length_ratios}
\end{table}
\section{Additional Results}
\label{appendix:b}

Table~\ref{tab:pretraining-all-metrics} is an extension of Table~\ref{tab:pretraining} in Section~\ref{sec:results_pretraining}, and shows pretraining results across all metrics (including \textit{BLEURT} and \textit{Comet20}) on the WMT'23 en$\rightarrow$de and de$\rightarrow$en test sets.
\begin{table*}[t]
    \centering
    \begin{adjustbox}{width=\textwidth}
\begin{tabular}{llcccc}
\toprule
& \bf Model & \bf BLEURT $\uparrow$ & \bf MetricX $\downarrow$ & \bf COMET20 $\uparrow$ & \bf COMET22 $\uparrow$ \\
\midrule
\multirow{6}{*}{\textit{en$\rightarrow$de}}
& 1a) WMT'23 (all) & 64.11 & 4.20 & 42.52 & 78.79 \\
& 1b) WMT'23 (sentence-level, \textit{BLEURT-QE} filtered) & 29.33 & 16.69 & -1.14 & 43.18 \\
\cmidrule{3-6}
& 2a) Greedy sentence-level + blob-level (9:1) & \bf 67.75 & \bf 2.60 & \bf 51.20 & \bf 81.67 \\
& 2b) MBR sentence-level & 56.93 & 6.39 & 21.86 & 72.05 \\
& 2c) QE blob-level & 67.18 & 2.82 & 45.72 & 80.62 \\
& 2d) MBR sentence-level + QE blob-level (9:1) & 66.34 & 2.99 & 43.20 & 79.68 \\
\midrule[1.5pt]
\multirow{6}{*}{\textit{de$\rightarrow$en}}
& 1a) WMT'23 (all) & 69.96 & 5.55 & 51.52 & 82.41 \\
& 1b) WMT'23 (sentence-level, \textit{BLEURT-QE} filtered) & 44.80 & 14.80 & -64.27 & 57.27 \\
\cmidrule{3-6}
& 2a) Greedy sentence-level + blob-level (9:1) & \bf 70.94 & \bf 3.47 & \bf 54.86 & \bf 83.30 \\
& 2b) MBR sentence-level & 67.21 & 4.97 & 44.12 & 80.55 \\
& 2c) QE blob-level & 69.68 & 4.01 & 51.00 & 82.33 \\
& 2d) MBR sentence-level + QE blob-level (9:1) & 69.07 & 3.95 & 49.71 & 82.02 \\
\bottomrule
\end{tabular}
\end{adjustbox}
\caption{Pretraining performance (WMT'23 test set).}
\label{tab:pretraining-all-metrics}
\end{table*}
Table~\ref{tab:greedyVsMBR} illustrates the stylistic differences between greedy and MBR decoding.
\begin{table*}
    \newlength{\exampleWidth}
    \setlength{\exampleWidth}{12cm}

    \centering

    \begin{tabular}{lp{\exampleWidth}}
    \toprule
    Source & "While President Trump has not commented on Lewis' death, White House press secretary Kayleigh McEnany said in a tweet Saturday, [\dots] \\
    \midrule
    Greedy & Während Präsident Trump sich nicht zu Lewis' Tod geäußert hat, sagte die Pressesprecherin des Weißen Hauses, Kayleigh McEnany, am Samstag in einem Tweet: [\dots] \\
    \midrule
    MBR & Präsident Trump hat bislang keinen Kommentar zu Lewis’ Tod abgegeben. Weißes Hauses-Pressesprecherin Kayleigh McEnany äußerte sich hingegen auf Twitter: [\dots] \\
    \midrule
    Explanation & Greedy decoding closely follows the English sentence structure. MBR decoding produces a more natural German sentence, and also decides to introduce a sentence break. \\
    \bottomrule
    \end{tabular}

    \vspace*{1em}

    \begin{tabular}{lp{\exampleWidth}}
    \toprule
    Source & "If we follow the Hungarian philosopher Gáspár Miklós Tamás, with his very broad definition of fascism as ""a break with the enlightenment tradition of citizenship as a universal entitlement,"" the similarities sharpen." \\
    \midrule
    Greedy & Wenn wir dem ungarischen Philosophen Gáspár Miklós Tamás mit seiner sehr weiten Definition von Faschismus als „Bruch mit der aufklärerischen Tradition der Staatsbürgerschaft als universellem Recht“ folgen, werden die Ähnlichkeiten deutlicher. \\
    \midrule
    MBR &  Wenn man dem ungarischen Philosophen Gàspár Miklós Tamás folgt, der in seiner sehr breiten Definition des Faschismus eine „Verletzung der aufgeklärten Tradition des Bürgerstatus als universalem Recht“ sieht, dann wird die Ähnlichkeit noch deutlicher. \\
    \midrule
    Explanation & The MBR translation uses an impersonal form, which is preferred in German.
    The subordinate clause about the definition of fascism is also reformulated, resulting in a more natural structure. \\
    \bottomrule
    \end{tabular}

    \caption{Comparison between translations generated with greedy and MBR decoding.}
    \label{tab:greedyVsMBR}
\end{table*}

Table~\ref{tab:finetuning-all-metrics} is an extension of Table~\ref{tab:finetuning} in Section~\ref{sec:results_finetuning}, and shows finetuning results on the WMT'23 test set across all metrics.
\begin{table*}[t]
    \centering
    \begin{adjustbox}{width=\textwidth}
\begin{tabular}{llcccc}
\toprule
& \bf Model & \bf BLEURT $\uparrow$ & \bf MetricX $\downarrow$ & \bf COMET20 $\uparrow$ & \bf COMET22 $\uparrow$ \\
\midrule
\multirow{6}{*}{\textit{en$\rightarrow$de}}
& 1a) Greedy sentence-level + blob-level (9:1) & 68.31 & 2.59 & 51.91 & 81.49 \\
& 1b) MBR sentence-level & \bf 70.65 & 2.30 & \bf 55.38 & \bf 82.69 \\
& 1c) QE blob-level & 68.19 & 2.45 & 51.97 & 81.83 \\
& 1d) MBR sentence-level + QE blob-level (9:1) & 70.35 & \bf 2.26 & 55.07 & 82.52 \\
\cmidrule{3-6}
& 2a) PaLM-2 five-shot (no finetuning) & 72.34 & 1.62 & 60.62 & 84.54 \\
& 2b) PaLM-2 MBR sentence-level & \bf 74.38 & \bf 1.14 & \bf 64.86 & \bf 85.64 \\
& 2c) PaLM-2 QE blob-level & 72.31 & 1.47 & 61.03 & 84.77 \\
& 2d) PaLM-2 MBR sentence-level + QE blob-level (9:1) & 74.21 & 1.17 & 64.43 & 85.54 \\
\midrule[1.5pt]
\multirow{6}{*}{\textit{de$\rightarrow$en}}
& 1a) Greedy sentence-level + blob-level (9:1) & 72.61 & 3.12 & 59.52 & 84.14 \\
& 1b) MBR sentence-level & \bf 73.56 & 2.91 & \bf 61.47 & \bf 84.57 \\
& 1c) QE blob-level & 73.02 & 2.99 & 59.73 & 84.27 \\
& 1d) MBR sentence-level + QE blob-level (9:1) & 73.47 & \bf 2.82 & 61.04 & 84.53 \\
\midrule
& 2a) PaLM-2 five-shot (no finetuning) & 74.73 & 2.25 & 64.72 & 85.36 \\
& 2b) PaLM-2 MBR sentence-level & \bf 76.20 & \bf 1.91 & \bf 68.41 & \bf 86.26 \\
& 2c) PaLM-2 QE blob-level & 75.56 & 2.03 & 66.55 & 85.81 \\
& 2d) PaLM-2 MBR sentence-level + QE blob-level (9:1) & 76.18 & 1.92 & 68.12 & 86.24 \\
\bottomrule
\end{tabular}
\end{adjustbox}
\caption{Finetuning performance (WMT'23 test set). Unless otherwise indicated, performance is reported for the encoder-decoder model. For finetuning, this model was initialized from the checkpoint pretrained on the full WMT'23 training dataset (row 1a in Table~\ref{tab:pretraining-all-metrics}). Results for \textit{PaLM-2 Bison} few-shot prompting versus self-distillation using \texttt{NewsPaLM} MBR and QE data are reported in rows 2a-d.}
\label{tab:finetuning-all-metrics}
\end{table*}
Table~\ref{tab:wmt24-ende} shows all en$\rightarrow$de pretraining and finetuning results on the WMT'24 test set. (Note that there does not exist a WMT'24 de$\rightarrow$en test set.) Table~\ref{tab:wmt24-mbr-greedy} is the companion to Table~\ref{tab:mbr-greedy} in Section~\ref{sec:results}, but on the WMT'24 (rather than WMT'23) test set.
\begin{table*}[t]
    \centering
    \begin{adjustbox}{width=\textwidth}
\begin{tabular}{lcccc}
\toprule
\bf Model & \bf BLEURT $\uparrow$ & \bf MetricX $\downarrow$ & \bf COMET20 $\uparrow$ & \bf COMET22 $\uparrow$ \\
\midrule
1a) PT: WMT'23 (all) & 65.08 & 3.15 & 29.18 & 77.79 \\
1b) PT: WMT'23 (sentence-level, Bleurt-QE filtered) & 34.62 & 13.06 & -90.56 & 49.38 \\
\midrule
1c) PT: Greedy sentence-level + blob-level (9:1) & 64.78 & 2.95 & 27.88 & 77.81 \\
1d) PT: MBR sentence-level & 55.82 & 5.32 & -0.20 & 69.88 \\
1e) PT: QE blob-level & 63.80 & 3.17 & 22.34 & 76.72 \\
1f) PT: MBR sentence-level + QE blob-level (9:1) & 64.27 & 3.13 & 21.32 & 75.93 \\
\midrule[1.5pt]
2a) FT: Greedy sentence-level + blob-level (9:1) & 67.94 & 2.26 & 39.22 & 80.20 \\
2b) FT: MBR sentence-level & 70.33 & 1.96 & 41.96 & 80.96 \\
2c) FT: QE blob-level & 68.14 & 2.27 & 37.07 & 79.88 \\
2d) FT: MBR sentence-level + QE blob-level (9:1) & 70.04 & 2.00 & 41.83 & 80.81 \\
\midrule
2e) FT: PaLM-2 five-shot (no finetuning) & 72.37 & 1.28 & 49.39 & 83.51 \\
2f) FT: PaLM-2 MBR sentence-level & 73.94 & 1.05 & 53.99 & 84.44 \\
2g) FT: PaLM-2 QE blob-level & 71.34 & 1.40 & 46.16 & 82.84 \\
2h) FT: PaLM-2 MBR sentence-level + QE blob-level (9:1) & 73.83 & 1.05 & 53.57 & 84.31 \\
\bottomrule
\end{tabular}
\end{adjustbox}
\caption{Pretraining and finetuning performance (en$\rightarrow$de WMT'24 test set). The PT prefix indicates pretrained models, and the FT prefix indicates finetuned models. Unless otherwise indicated, performance is reported for the encoder-decoder model. For finetuning, this model was initialized from the checkpoint pretrained on the full WMT'23 training dataset (row 1a). Results for \textit{PaLM-2 Bison} few-shot prompting versus self-distillation using \texttt{NewsPaLM} MBR and QE data are reported in rows 2e-h.}
\label{tab:wmt24-ende}
\end{table*}

\begin{table*}[t]
    \centering
    \begin{adjustbox}{width=\textwidth}
\begin{tabular}{lcccc}
\toprule
\bf Model & \bf BLEURT $\uparrow$ & \bf MetricX $\downarrow$ & \bf COMET20 $\uparrow$ & \bf COMET22 $\uparrow$ \\
\midrule
MBR + QE finetuning (from greedy-pretrained ckpt) & \bf 68.12 & \bf 2.41 & \bf 33.75 & \bf 79.37 \\
Greedy finetuning (from MBR + QE-pretrained ckpt) & 65.61 & 2.91 & 27.50 & 77.66 \\
\bottomrule
\end{tabular}
\end{adjustbox}
\caption{Comparison of pretraining on \texttt{NewsPaLM} greedy data, then finetuning on \texttt{NewsPaLM} MBR and QE data, versus vice-versa (en$\rightarrow$de WMT'24 test set).}
\label{tab:wmt24-mbr-greedy}
\end{table*}
Tables~\ref{tab:wmt23-ende-per-domain} and \ref{tab:wmt24-ende-per-domain} show the pretraining and finetuning results on the en$\rightarrow$de WMT'23 and WMT'24 test sets, respectively, broken out by domain. For WMT'23, the domains are \textit{Mastodon}, \textit{News}, \textit{Speech}, and \textit{User Review}. For WMT'24, the domains are \textit{Literary}, \textit{News}, \textit{Social}, and \textit{Speech}. Note that the models pretrained and finetuned on our \texttt{NewsPaLM} dataset perform especially strongly on the \textit{News} domain, but the gains aren't limited to this domain.
\begin{table*}[t]
    \centering
    \begin{adjustbox}{width=\textwidth}
\begin{tabular}{lcccccccc}
\toprule
& \multicolumn{2}{c}{\bf Mastodon} & \multicolumn{2}{c}{\bf News} & \multicolumn{2}{c}{\bf Speech} & \multicolumn{2}{c}{\bf User Review} \\
\cmidrule{2-9}
\bf Model & \bf MetricX $\downarrow$ & \bf COMET22 $\uparrow$ & \bf MetricX $\downarrow$ & \bf COMET22 $\uparrow$ & \bf MetricX $\downarrow$ & \bf COMET22 $\uparrow$ & \bf MetricX $\downarrow$ & \bf COMET22 $\uparrow$ \\
\midrule
1a) PT: WMT'23 (all) & 4.09 & 79.26 & 3.97 & 80.46 & 3.87 & 78.55 & 5.19 & 75.14 \\
\midrule
1b) PT: Greedy sentence-level + blob-level (9:1) & 2.33 & 81.98 & 1.59 & 84.72 & 3.44 & 79.30 & 3.74 & 79.10 \\
1c) PT: MBR sentence-level & 4.83 & 72.22 & 4.41 & 77.98 & 8.09 & 70.78 & 10.88 & 64.17 \\
1d) PT: QE blob-level & 2.70 & 80.27 & 1.75 & 84.96 & 3.63 & 77.37 & 3.71 & 78.68 \\
1e) PT: MBR sentence-level + QE blob-level (9:1) & 2.73 & 79.60 & 1.79 & 84.28 & 3.82 & 77.90 & 4.39 & 75.02 \\
\midrule[1.5pt]
2a) FT: Greedy sentence-level + blob-level (9:1) & 2.50 & 81.32 & 2.05 & 83.73 & 2.94 & 80.40 & 3.23 & 79.69 \\
2b) FT: MBR sentence-level & 2.03 & 82.87 & 1.87 & 84.94 & 2.75 & 80.75 & 3.01 & 81.15 \\
2c) FT: QE blob-level & 2.36 & 81.10 & 1.86 & 84.42 & 2.72 & 80.44 & 3.21 & 81.15 \\
2d) FT: MBR sentence-level + QE blob-level (9:1) & 2.17 & 82.21 & 1.76 & 84.91 & 2.54 & 80.94 & 2.87 & 81.40 \\
\midrule
2e) FT: PaLM-2 five-shot (no finetuning) & 1.40 & 84.86 & 1.15 & 86.11 & 2.45 & 82.47 & 1.83 & 83.82 \\
2f) FT: PaLM-2 MBR sentence-level & 0.97 & 86.00 & 0.88 & 86.48 & 1.70 & 83.33 & 1.22 & 86.21 \\
\bottomrule
\end{tabular}
\end{adjustbox}
\caption{Per-domain results on en$\rightarrow$de WMT'23 test set. The PT prefix indicates pretrained models, and the FT prefix indicates finetuned models. Unless otherwise indicated, performance is reported for the encoder-decoder model. For finetuning, this model was initialized from the checkpoint pretrained on the full WMT'23 training dataset (row 1a). Results for \textit{PaLM-2 Bison} few-shot prompting versus self-distillation using \texttt{NewsPaLM} MBR data are reported in rows 2e-f.}
\label{tab:wmt23-ende-per-domain}
\end{table*}

\begin{table*}[t]
    \centering
    \begin{adjustbox}{width=\textwidth}
\begin{tabular}{lcccccccc}
\toprule
& \multicolumn{2}{c}{\bf Literary} & \multicolumn{2}{c}{\bf News} & \multicolumn{2}{c}{\bf Social} & \multicolumn{2}{c}{\bf Speech} \\
\cmidrule{2-9}
\bf Model & \bf MetricX $\downarrow$ & \bf COMET22 $\uparrow$ & \bf MetricX $\downarrow$ & \bf COMET22 $\uparrow$ & \bf MetricX $\downarrow$ & \bf COMET22 $\uparrow$ & \bf MetricX $\downarrow$ & \bf COMET22 $\uparrow$ \\
\midrule
1a) PT: WMT'23 (all) & 3.79 & 75.95 & 2.86 & 81.20 & 2.78 & 76.44 & 3.67 & 80.04 \\
\midrule
1b) PT: Greedy sentence-level + blob-level (9:1) & 6.11 & 68.47 & 1.40 & 84.42 & 2.44 & 77.36 & 2.22 & 82.80 \\
1c) PT: MBR sentence-level & 9.79 & 59.40 & 3.39 & 79.32 & 4.44 & 68.81 & 4.61 & 75.44 \\
1d) PT: QE blob-level & 6.53 & 68.01 & 1.38 & 84.01 & 2.69 & 75.95 & 2.39 & 81.38 \\
1e) PT: MBR sentence-level + QE blob-level (9:1) & 6.11 & 68.47 & 1.40 & 84.42 & 2.44 & 77.36 & 2.22 & 82.80 \\
\midrule[1.5pt]
2a) FT: Greedy sentence-level + blob-level (9:1) & 3.12 & 77.89 & 1.64 & 84.00 & 2.13 & 78.85 & 2.18 & 82.69 \\
2b) FT: MBR sentence-level & 2.83 & 78.56 & 1.40 & 84.78 & 1.79 & 79.70 & 1.96 & 83.36 \\
2c) FT: QE blob-level & 3.13 & 77.38 & 1.75 & 84.13 & 2.10 & 78.42 & 2.26 & 82.51 \\
2d) FT: MBR sentence-level + QE blob-level (9:1) & 2.91 & 78.26 & 1.36 & 84.88 & 1.85 & 79.56 & 1.96 & 83.11 \\
\midrule
2e) FT: PaLM-2 five-shot (no finetuning) & 1.42 & 82.24 & 1.08 & 84.89 & 1.23 & 82.79 & 1.38 & 85.25 \\
2f) FT: PaLM-2 MBR sentence-level & 1.24 & 83.55 & 0.86 & 85.79 & 0.99 & 83.76 & 1.09 & 85.75 \\
\bottomrule
\end{tabular}
\end{adjustbox}
\caption{Per-domain results on en$\rightarrow$de WMT'24 test set. The PT prefix indicates pretrained models, and the FT prefix indicates finetuned models. Unless otherwise indicated, performance is reported for the encoder-decoder model. For finetuning, this model was initialized from the checkpoint pretrained on the full WMT'23 training dataset (row 1a). Results for \textit{PaLM-2 Bison} few-shot prompting versus self-distillation using \texttt{NewsPaLM} MBR data are reported in rows 2e-f.}
\label{tab:wmt24-ende-per-domain}
\end{table*}

\begin{figure}[!htp]
    \centering
    \includegraphics[width=\columnwidth]{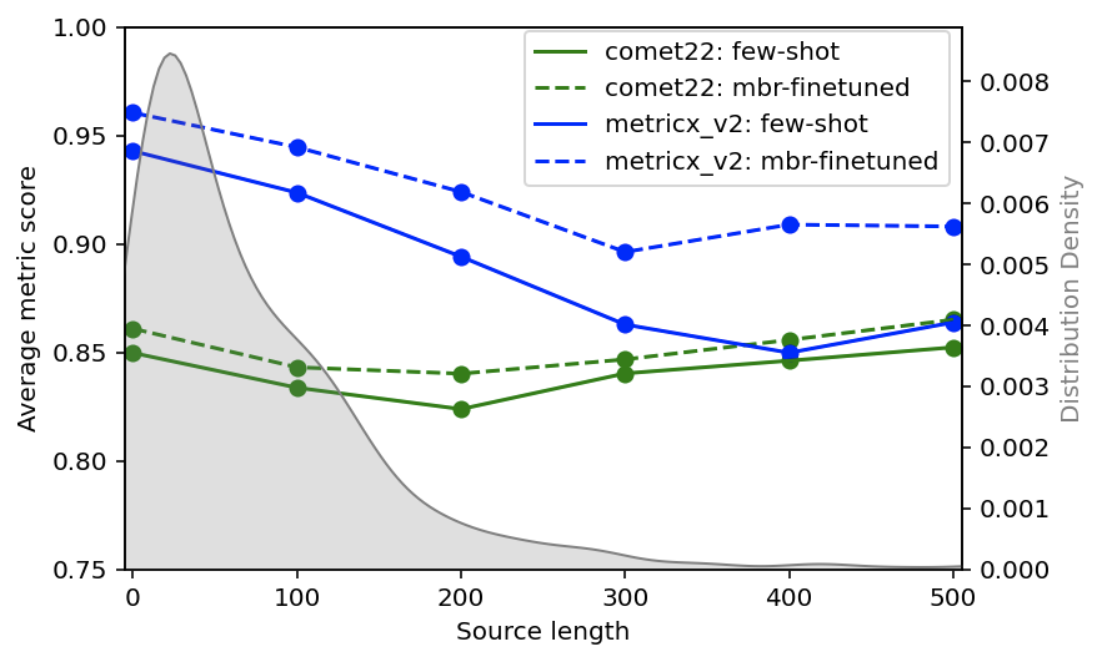}
    \caption{\textit{PaLM-2 Bison} few-shot versus \texttt{NewsPaLM} MBR-finetuned performance bucketed by source length (en$\rightarrow$de WMT'23 test set). Note that self-MBR finetuning (on sentence-level data only) improves performance across all source length buckets.}
    \label{fig:results_ulm_bucketed}
\end{figure}

Figure~\ref{fig:results_ulm_bucketed} shows the \textit{PaLM-2 Bison} few-shot versus self-MBR-finetuned results on the en$\rightarrow$de WMT'23 test set, bucketed by source segment length. Note that the gains in performance from self-distillation are consistent across all segment length buckets.

Table~\ref{tab:subsampling} accompanies Figure~\ref{fig:subsampling} in Section~\ref{sec:ablation_scaling}, and shows the results of the \texttt{NewsPaLM} subsampling ablations across all metrics.
\begin{table*}[t]
    \centering
    \begin{adjustbox}{width=\textwidth}
\begin{tabular}{llcccc}
\toprule
& \bf Model & \bf BLEURT $\uparrow$ & \bf MetricX $\downarrow$ & \bf COMET20 $\uparrow$ & \bf COMET22 $\uparrow$ \\
\midrule
\multirow{4}{*}{\textit{en$\rightarrow$de}}
& 1a) PT: Full dataset & 56.93 & 6.39 & 21.86 & 72.05 \\
& 1b) PT: Subsampled dataset & 43.73 & 11.02 & -25.72 & 59.75 \\
\cmidrule{2-6}
& 2a) FT: Full dataset & 70.65 & 2.30 & 55.38 & 82.69 \\
& 2b) FT:  Subsampled dataset & 70.00 & 2.55 & 53.91 & 82.11 \\
\midrule[1.5 pt]
\multirow{4}{*}{\textit{de$\rightarrow$en}}
& 1a) PT: Full dataset & 67.21 & 4.97 & 44.12 & 80.55 \\
& 1b) PT: Subsampled dataset & 60.43 & 7.41 & 22.89 & 75.91 \\
\cmidrule{2-6}
& 2a) FT: Full dataset & 73.56 & 2.91 & 61.47 & 84.57 \\
& 2b) FT:  Subsampled dataset & 73.15 & 2.96 & 59.35 & 84.33 \\
\bottomrule
\end{tabular}
\end{adjustbox}
\caption{Comparison of model performance when pretraining and finetuning on the full versus subsampled \texttt{NewsPaLM} MBR dataset (WMT'23 test set). The subsampled dataset is 25\% of the size of the full dataset, and was sampled randomly. The PT prefix indicates pretrained models, and the FT prefix indicates finetuned models.}
\label{tab:subsampling}
\end{table*}

\end{document}